\documentclass[letterpaper]{article} 
\usepackage{aaai2026}  
\usepackage{times}  
\usepackage{helvet}  
\usepackage{courier}  
\usepackage[hyphens]{url}  
\usepackage{graphicx} 
\usepackage{natbib}  
\usepackage{caption} 
\usepackage{algorithm}
\usepackage{algorithmic}
\usepackage{amsfonts}
\usepackage{amsmath}
\usepackage{booktabs}

\usepackage{multirow}
\usepackage{array}
\usepackage{xcolor}
\usepackage{colortbl}
\definecolor{lightblue}{RGB}{220, 235, 255}
\definecolor{lightgreen}{RGB}{220, 255, 235}
\usepackage{newfloat}
\usepackage{listings}
\DeclareCaptionStyle{ruled}{labelfont=normalfont,labelsep=colon,strut=off} 
\floatstyle{ruled}
\newfloat{listing}{tb}{lst}{}
\floatname{listing}{Listing}
\usepackage{hyperref}
\nocopyright

\usepackage{xcolor}
\usepackage{xspace}
\usepackage{gradient-text}

\title{%
    \includegraphics[height=2.5em]{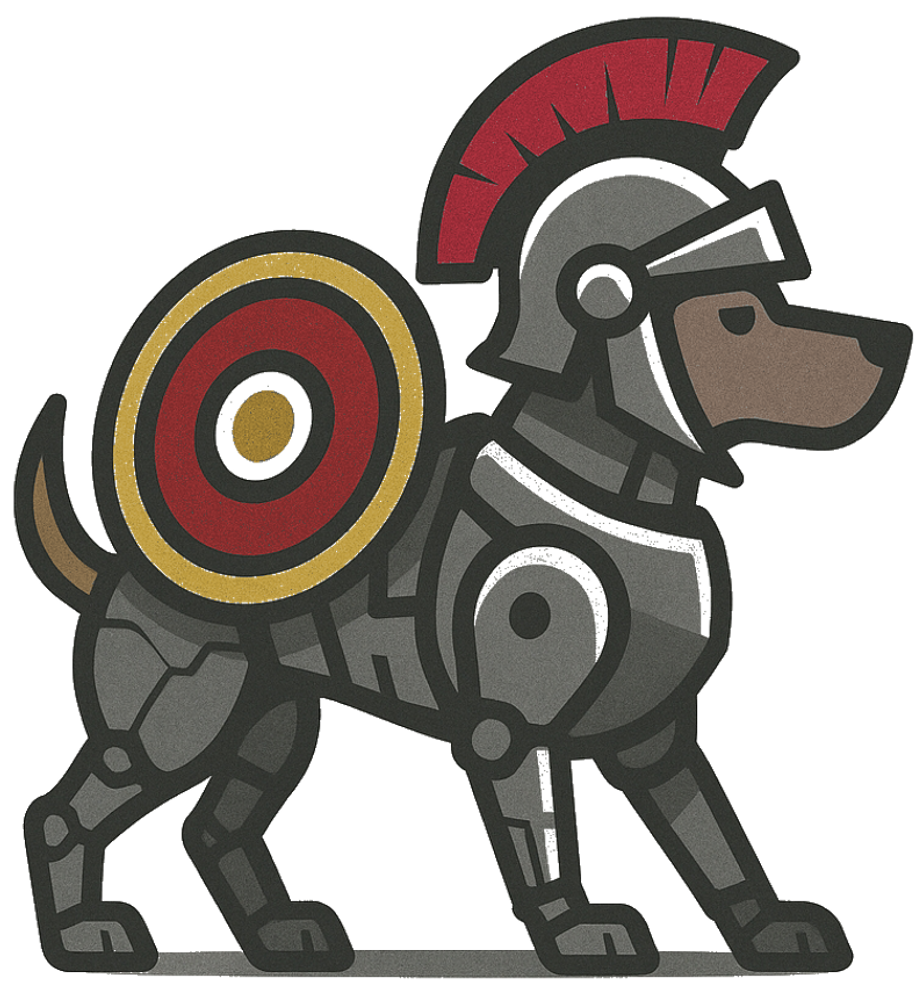} 
    ODYSSEY: Open-World Quadrupeds \\ 
    Exploration and Manipulation for Long-Horizon Tasks%
}
\author{
    Kaijun Wang\textsuperscript{\rm 1}\equalcontrib,
    Liqin Lu\textsuperscript{\rm 2}\equalcontrib,
    Mingyu Liu\textsuperscript{\rm 1},
    Jianuo Jiang\textsuperscript{\rm 3},
    Zeju Li\textsuperscript{\rm 1},
    Bolin Zhang\textsuperscript{\rm 1},
    Wancai Zheng\textsuperscript{\rm 2},
    Xinyi Yu\textsuperscript{\rm 2}\thanks{Corresponding authors.},
    Hao Chen\textsuperscript{\rm 1}\footnotemark[2],
    Chunhua Shen\textsuperscript{\rm 1}\footnotemark[2]
}
\affiliations{
    \textsuperscript{\rm 1}Zhejiang University\\
    \textsuperscript{\rm 2}Zhejiang University of Technology\\
    \textsuperscript{\rm 3}The Chinese University of Hong Kong, Shenzhen\\

}

\usepackage{bibentry}

\begin{document}

\makeatletter
\let\@oldmaketitle\@maketitle
\renewcommand{\@maketitle}{
   \@oldmaketitle
 \begin{center}
   \vspace{-5ex}
      \includegraphics[width=0.95\linewidth]{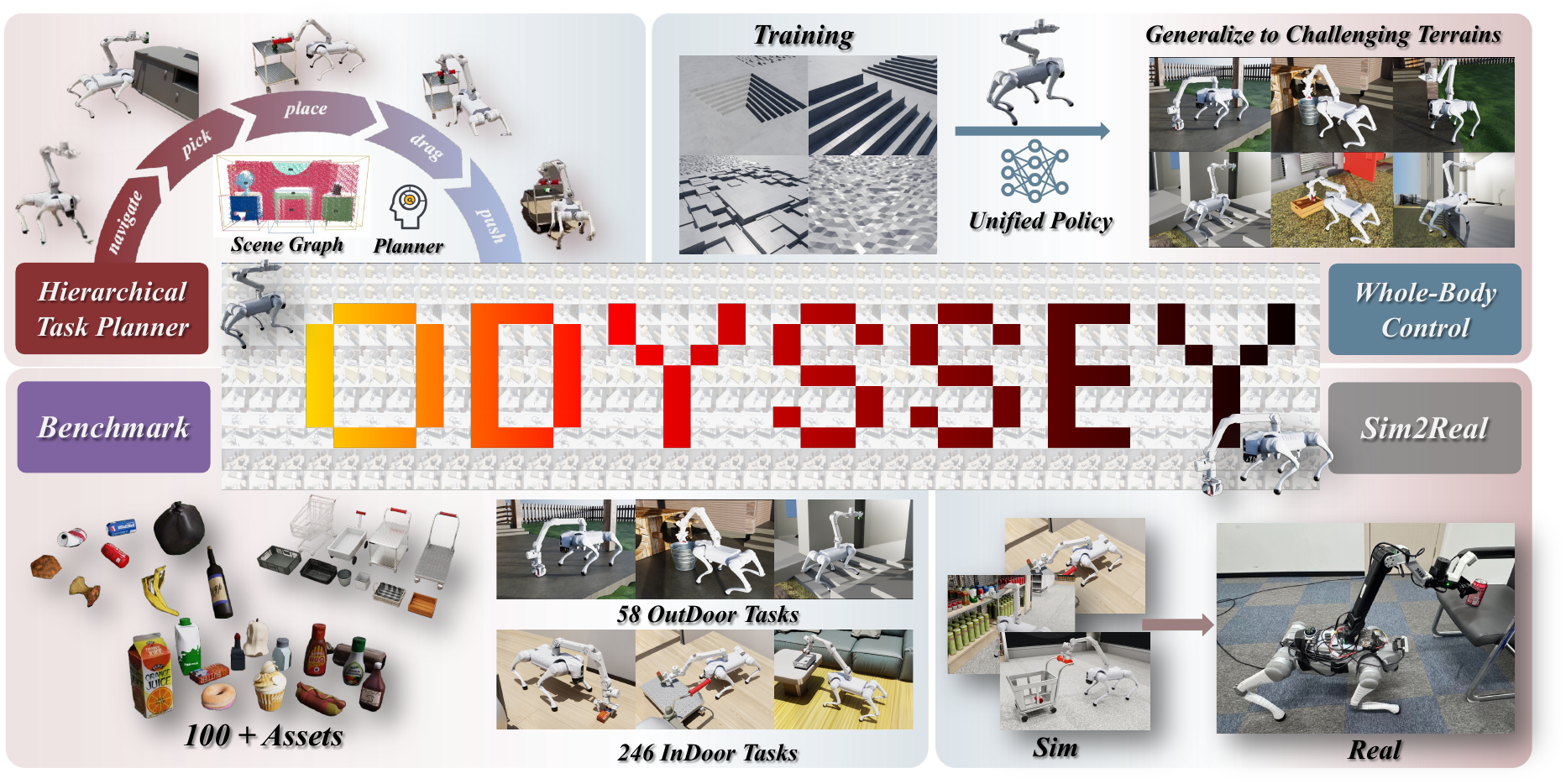}
 \end{center}
 
  \refstepcounter{figure}\normalfont Figure~\thefigure. 
  We present ODYSSEY, a unified mobile manipulation framework for agile quadruped robots equipped with
manipulators, which seamlessly integrates high-level task
planning with low-level whole-body control. 
  \label{fig:teaser}
  \newline
  }
\maketitle


\begin{abstract}

Language-guided long-horizon mobile manipulation has long been a grand challenge in embodied semantic reasoning, generalizable manipulation, and adaptive locomotion. Three fundamental limitations hinder progress: First, although large language models have shown promise in enhancing spatial reasoning and task planning through learned semantic priors, existing implementations remain confined to tabletop scenarios, failing to address the constrained perception and limited actuation ranges characteristic of mobile platforms. Second, current manipulation strategies exhibit insufficient generalization when confronted with the diverse object configurations encountered in open-world environments. Third, while crucial for practical deployment, the dual requirement of maintaining high platform maneuverability alongside precise end-effector control in unstructured settings remains understudied in the literature.

In this work, we present ODYSSEY, a unified mobile manipulation framework for agile quadruped robots equipped with manipulators, which seamlessly integrates high-level task planning with low-level whole-body control. To address the challenge of egocentric perception in language-conditioned tasks, we introduce a hierarchical planner powered by a vision-language model, enabling long-horizon instruction decomposition and precise action execution. At the control level, our novel whole-body policy achieves robust coordination of locomotion and manipulation across challenging terrains. We further present the first comprehensive benchmark for long-horizon mobile manipulation, evaluating diverse indoor and outdoor scenarios. Through successful sim-to-real transfer, we demonstrate the system’s generalization and robustness in real-world deployments, underscoring the practicality of legged manipulators in unstructured environments. Our work advances the feasibility of generalized robotic assistants capable of complex, dynamic tasks. Our project page: \href{https://kaijwang.github.io/odyssey.github.io/}{https://kaijwang.github.io/odyssey.github.io/}


\end{abstract}


\section{Introduction}

Open-world mobile manipulation allows robots to autonomously navigate and interact in dynamic, unstructured environments by tightly integrating mobility, manipulation, and real-time perception. Unlike traditional methods that separate navigation and manipulation, this unified approach unlocks emergent capabilities like active perception, which is essential for real-world tasks. For example, a robot might subtly adjust its position while reaching for an object to gain a better grasp position. This adaptive behavior emerges naturally in complex environments where static perception or sequential planning would fail.

Prior research has achieved robust solutions for navigation in dynamic environments~\cite{grandia2023perceptive, zhuang2023robot, liu2024dynamem} and manipulation in controlled settings~\cite{kim2024openvla,brohan2022rt,cheang2024gr}. While recent works~\cite{pan2025roboduet, fu2023deep,liu2024visual, zhang2025slim, wang2024quadwbg} have developed whole-body control frameworks, they face scalability limitations in open-world scenarios due to simplified environmental assumptions and evaluations restricted to short-horizon tasks. We present ODYSSEY, a reinforcement learning-based whole-body control system that unifies robust quadruped locomotion with precise manipulation through an integrated vision-language framework. Our approach achieves state-of-the-art control accuracy under out-of-distribution conditions while significantly expanding operational capabilities. Unlike prior work, we demonstrate generalization across diverse challenging terrains, enabling real-world deployment.

Recent works~\cite{qi2025sofar, pan2025omnimanip} demonstrate that large language models can substantially enhance robotic task planning and generalizable manipulation through their spatial understanding abilities. We extend the capability of large language models to whole-body navigation and manipulation tasks. Specifically, our approach grounds task execution at two levels: task-level planning over a semantic instance map and fine-grained action guidance via geometry-constrained pose estimation.

Besides, to address the critical evaluation gap, we present the first comprehensive benchmark for evaluating long-horizon mobile manipulation, featuring eight diverse daily tasks across indoor/outdoor environments with hundreds of object configurations. The benchmark enables holistic assessment of embodied reasoning, task planning, navigation, and manipulation capabilities, while incorporating the standardized Arnold framework for precise manipulation evaluation.

Through extensive experiments, our system shows strong ability for sim2real transfer, showcasing exceptional generalization where both control and planning modules maintain consistent performance across diverse real-world scenarios. Our contributions are four folds:

(i) We introduce a hierarchical vision-language planner that bridges egocentric perception and language-conditioned tasks, decomposing long-horizon instructions into executable actions.

(ii) We propose the first whole-body control policy that generalizes to challenging terrains while jointly coordinating locomobility and manipulation.

(iii) We introduce the first long-horizon mobile manipulation benchmark, covering a wide range of realistic indoor and outdoor scenarios.

(iv) We further demonstrate successful sim-to-real transfer of both high-level planners and low-level control policies, showing strong generalization and robustness in real-world deployments. 

Our results highlight the feasibility and practicality of deploying a legged mobile manipulator in unstructured environments, paving the way toward generalized robotic assistants.

\section{Related Work}

\subsection{Open-world mobile manipulation}
Previous research has made significant advances in both navigation and manipulation as separate domains, with robust solutions developed for mobile robot path planning in dynamic environments~\cite{grandia2023perceptive, zhuang2023robot} and sophisticated manipulation techniques for object interaction in controlled settings~\cite{kim2024openvla,brohan2022rt,cheang2024gr}. While pioneering works~\cite{pan2025roboduet, fu2023deep,liu2024visual, zhang2025slim, wang2024quadwbg, fu2024mobile, jiang2025behavior} have developed initial whole-body control frameworks, they face scalability limitations in open-world scenarios due to oversimplified environmental assumptions and evaluations limited to short-horizon pick-and-place tasks. Some works~\cite{ha2024umi, qiu2024wildlma} have attempted to attach more complex interactions by learning per-action policies from human demonstrations. However, these methods lack compositionality and scalability, needing task-specific data for each scenario. ODYSSEY overcomes these limitations by unifying terrain-aware locomotion with hierarchical planning, enabling robust mobile manipulation in unstructured environments.


\subsection{Foundation Models for Embodied Tasks}
Vision-language models (VLMs) have shown promise in enhancing robotic reasoning ~\cite{qi2025sofar, pan2025omnimanip, zhi2024closed, qiu2024learning, wang2025vq}, but their evaluation has been restricted to tabletop settings with fixed cameras. For navigation, foundation models improve spatial understanding ~\cite{gu2024conceptgraphs, jatavallabhula2023conceptfusion, jiang2025dualmap}, yet they lack fine-grained manipulation support. ODYSSEY advances this by grounding hierarchical planning in egocentric perception, using VLMs to decompose tasks via scene graphs while generating precise end-effector trajectories. This contrasts with modular approaches ~\cite{zhang2025slim} that struggle with compositional reasoning under uncertainty.

\subsection{Benchmarks for Real-World Deployment}
Existing benchmarks for mobile manipulation ~\cite{qiu2024learning} focus narrowly on navigation or short-term interactions, lacking standardized metrics for long-horizon tasks. Simulation frameworks like IsaacSim have enabled progress in locomotion~\cite{zhi2024closed}, but manipulation-centric evaluations remain sparse. ODYSSEY introduces a comprehensive benchmark with diverse indoor/outdoor scenarios, addressing multi-stage reasoning, object-aware navigation, and precision manipulation. This complements prior work ~\cite{qiu2024wildlma} by enabling scalable testing of sim-to-real transfer for integrated control and planning.

\begin{figure*}[h]
\centering
\includegraphics[width=1.0\textwidth]{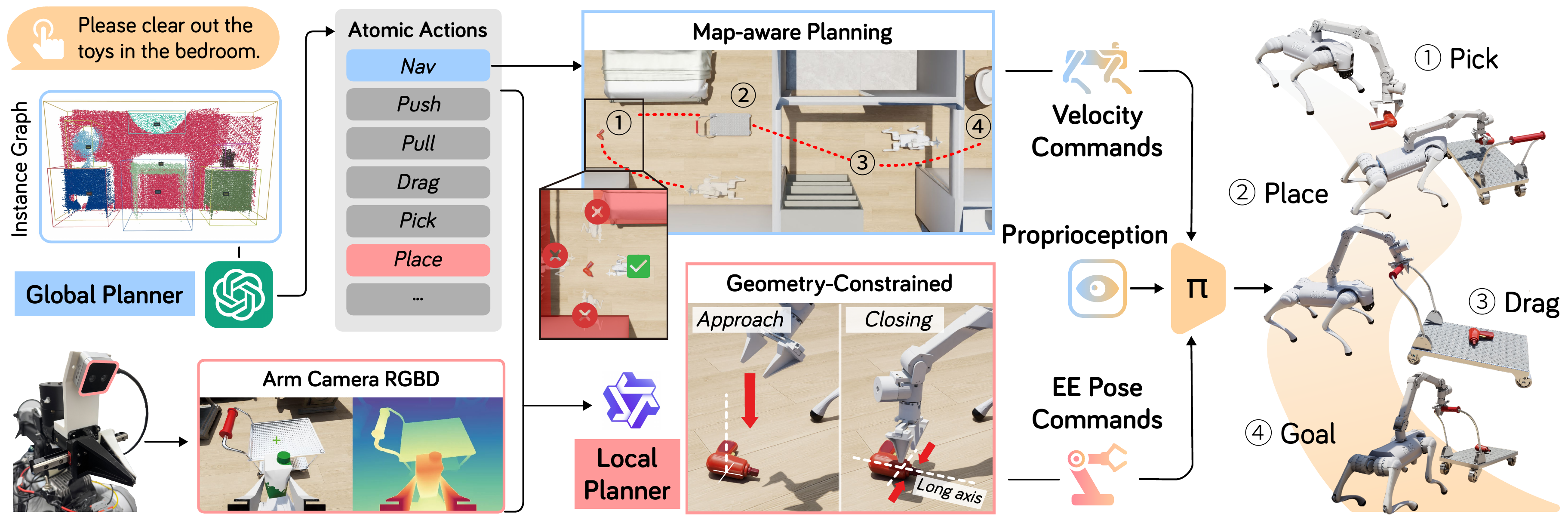}
\caption{ODYSSEY pipeline spans the entire process of a long-horizon task, including multi-modal semantic perception, map-aware global planning, geometry-constrained action grounding, and step-wise execution by a learned low-level policy.}
\label{fig: method_overall}
\end{figure*}

\section{Method}

In this section, we present ODYSSEY, a unified framework that spans long-horizon task planning, whole-body control, and standardized evaluation for mobile manipulation. It comprises three key components:

\begin{enumerate}
    \item \textbf{Coarse-to-Fine Task Planner} (Section~\ref{sec:planner}):  a hierarchical planner that orchestrates top-down task execution under the guidance of foundation models.

    \item \textbf{Quadruped Whole-Body Policy} (Section~\ref{sec:policy}): a reinforcement learning-based whole-body controller that generalizes to diverse terrains and overcomes sim-to-real gap.

    \item \textbf{Mobile Manipulation Benchmark} (Section~\ref{sec:bench}): the first scalable evaluation suite for assessing long-term task performance across versatile real-world scenarios.
\end{enumerate}

\subsection{Long-horizon Task Planner}
\label{sec:planner}

To bridge the gap left by prior work in modeling the intricate dependencies between semantic reasoning-based navigation and fine-grained, generalizable manipulation, our hierarchical framework is explicitly designed to ensure the reliability of both components while reinforcing their mutual dependencies for coherent long-horizon task execution.

\subsubsection{Map-Aware Task-Level Planning}
To support long-horizon task planning grounded in egocentric observations, we first build a global planner that integrates a lightweight multi-modal perception module as a plug-in component. Concretely, we fuse on-board RGB and LiDAR streams to form a unified spatial-semantic representation of the scene. Leveraging a suite of pre-trained foundation models, we map an instance graph that encodes object geometry and semantics for symbolic task reasoning. The pipeline of this module is detailed in Appendix~\ref{app: graph constructor}.

As illustrated in Fig.~\ref{fig: method_overall}, given the instance-level semantic map, GPT-4.1~\cite{achiam2023gpt} is used to break down template-free natural language instructions into a sequence of atomic actions from a predefined set: \texttt{navigate}, \texttt{pick}, \texttt{place}, and \texttt{push/pull}/\texttt{drag}. Each action is paired with a language description that tracks task progress and provides guidance for local planning.

For actions involving spatial displacement (\texttt{navigate}, \texttt{drag}), the model is further prompted to output a coarse target waypoint to guide planning. We project this target onto a 2D occupancy map built via online SLAM from accumulated LiDAR scans. A local search is then performed around the projected waypoint to identify a collision-free goal pose, avoiding object bounding boxes and structural obstacles. This process yields a globally grounded task plan that aligns with the scene context and is feasible under physical constraints.

\subsubsection{Geometry-Constrained Local Manipulation}

For atomic actions requiring close-range manipulation, we use wrist-mounted depth observations to guide a vision-language model for precise end-effector pose generation. Despite the diverse physical nature of different actions, we unify their execution through a single visuomotor interface, eliminating the need for per-action heuristics.

Specifically, given an RGB observation and the corresponding textual description of the current atomic action, we employ Qwen2.5-VL-72B-Instruct~\cite{bai2025qwen2}, a model enhanced with pixel-level grounding capabilities, to infer a task-relevant contact point $p^* \in \mathbb{R}^2$ in the image space.

The contact point is projected onto the aligned depth image to recover its corresponding 3D position in the robot coordinate frame, denoted as $\mathbf{P}_{ee} \in \mathbb{R}^3$. We further prompt the model to generate the orientation $\mathbf{R}_{ee} \in SO(3)$ of the end-effector by determining the gripper's closing direction($x$-axis) and approaching direction($z$-axis), subject to the following geometric constraints:


\begin{itemize}
    \item \textbf{Axis-alignment constraint:} When the target object or contact region exhibits a dominant axis $\mathbf{a} \in \mathbb{R}^3$, both the $x$-axis and $z$-axis of the end-effector should be orthogonal to it:
    \begin{align}
        \mathbf{r}_x^\top \mathbf{a} = 0, \quad \mathbf{r}_z^\top \mathbf{a} = 0.
    \end{align}
    
    \item \textbf{Surface-normal constraint:} If the object is attached to a planar surface with normal vector $\mathbf{n} \in \mathbb{R}^3$, then the $z$-axis of the end-effector should align with the surface normal without violating the first constraint:
    \begin{align}
        \mathbf{r}_z \parallel \mathbf{n}, \quad \text{s.t.} \quad \mathbf{r}_z^\top \mathbf{a} = 0.
    \end{align}
\end{itemize}

By leveraging the expressive grounding capacity of Qwen-VL and constraining the output pose with interpretable geometric conditions, our system achieves reliable local guidance for interaction-intensive manipulation primitives. To the best of our knowledge, this constitutes the first fine-grained manipulation planning system without third-person observation or scripted policies, marking a significant step toward scalable deployment in mobile, in-the-wild environments.

\subsection{Policy for Whole-body Control}
\label{sec:policy}

To effectively execute commands from a high-level planner and adapt to diverse terrains, a whole-body control policy is essential. This work proposes a two-stage, learning-based policy that utilizes a neural network to generate desired joint positions from a set of observations. To enhance the policy's robustness, the training process incorporates a carefully designed, terrain-invariant end-effector sampling strategy and comprehensive domain randomization. The resulting controller is resilient to varied environmental interactions and can be directly deployed on a physical robot. This section first defines the policy and subsequently discusses the training methodology.

\begin{figure}[h]
    \centering
    \includegraphics[width=1.0\linewidth]{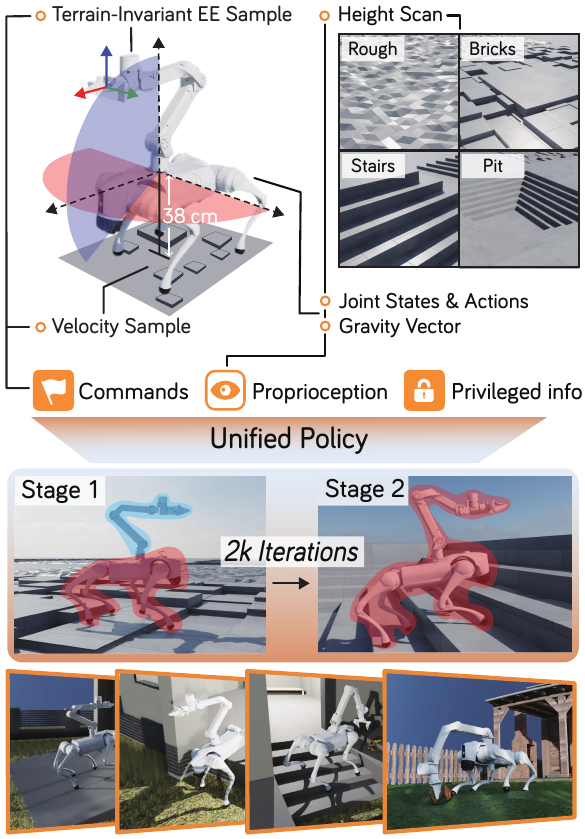} 
    \caption{An overview of the mobile manipulator policy and its two-stage training framework.}
    \label{fig: policy}
\end{figure}

\subsubsection{Mobile Manipulation Policy} \label{sec: policy define}
The mobile manipulation policy $\pi$ is formulated as a single network that maps a comprehensive observation vector to a target action $\mathbf{a}_t \in \mathbb{R}^{18}$ as shown in Eq.\ref{eq: policy}. The observation includes the locomotive command $\textbf{c}_t = (\hat{x}, \hat{y}, \hat{\omega})$, the 6-D end-effect target $\textbf{e}_t$, a local ground height map $\mathbf{m}_t$, the projected gravity vector $\mathbf{g}_{t}$, the previous timestep $\textbf{a}_{t-1} \in \mathbb{R}^{18}$ and the proprioceptive state $\mathbf{s}_t \in \mathbb{R}^{36}$ (joint positions $\mathbf{q}_t$ and velocities $\mathbf{\dot{q}}_t$). All commands and targets are expressed in the robot's base frame. To stabilize the policy output and reduce the simulation-to-reality gap~\cite{fu2023deep}, the action $\textbf{a}_t$ is formulated as the offsets to the default joint configuration ${\textbf{q}^{default} \in \mathbb{R}^{18} }$. The final target, ${\textbf{q}_t^{target} = \textbf{q}^{default} + \textbf{a}_t}$, is then converted to torques by a Proportional-Derivative (PD) controller.
\begin{align}
    \mathbf{s}_t &= (\mathbf{q}_t, \mathbf{\dot{q}})  \\
    \mathbf{a}_t &= \pi(\mathbf{c}_t, \mathbf{e}_t, \mathbf{s}_t, \mathbf{g}_{t}, \mathbf{m}_t, \mathbf{a}_{t-1})  \label{eq: policy}
\end{align}
To enhance training robustness and avoid local minima associated with searching a large action space, we employ a two-stage curriculum learning approach, as shown in Fig.~\ref{fig: policy}.

\noindent\textbf{Stage 1} 
In this stage, the arm joints are fixed to focus training on locomotion under a static load, improving exploration efficiency. Inspired by ~\cite{mittal2023orbit}, we introduce a gait reward incorporated alongside the base tracking reward to structure the robot's gait. Furthermore, a novel frequency reward is introduced to regulate the gait's cadence. The gait reward $r_{gait}$ encourages specific synchronous (e.g., diagonal) and asynchronous (e.g., lateral) foot contact patterns, with reward functions $r_s$ and $r_a$ detailed in the Appendix~\ref{app: reward design}.
\begin{align}
r_{gait} = \prod_{i,j \in \text{sync pairs}} r_{s}(i, j) \cdot \prod_{k,l \in \text{async pairs}} r_{a}(k, l)
\end{align}
The frequency reward $r_{fre}$ regulates the gait's cadence based on the error from a target frequency $f_{target}$. The gait frequency $f(leg)$ is the inverse of the time between consecutive ground contacts $(t_k^{cont} - t_{k-1}^{cont})$. The reward is then:
\begin{align}
    err(\text{leg}) &= (f(\text{leg}) - f_{target})^2  \\
    r_{f}(\text{leg}) &= \exp(-0.5 \cdot err(\text{leg})) \\
    r_{fre} &= \prod_{\text{leg} \in {\text{FL,FR,RL,RR}}} r_f(\text{leg})
\end{align}

\noindent\textbf{Stage 2} Following 2k training iterations, the process transitions to the second stage. In this stage, the policy controls all 18 joints, encompassing both the manipulator and the four legs. Consequently, the reward function is expanded to include end-effector tracking terms, $r_{arm}$, in addition to the previously described locomotion rewards, to guide the policy's training.

\noindent\textbf{Terrain-Invariant End-Effector Sampling} \label{sec:ee sample}
To ensure robust performance across varied topographies, our method employs a terrain-invariant end-effector sampling strategy. The process begins by sampling a target position from a spherical volume defined in the world coordinate system, centered at the robot's arm base. A key aspect of this strategy is that the target's z-axis height is fixed within the world frame before the coordinates are transformed into a Cartesian target position relative to the robot's moving base frame. This approach offers a significant advantage over sampling directly in the arm's local frame, as it effectively decouples the end-effector target from disturbances caused by changes in the robot's base pitch or the underlying terrain height. Consequently, this decoupling improves interaction accuracy during task execution.

\noindent\textbf{Domain Randomizatoin}
To bridge the simulation-to-reality gap, domain randomization is employed throughout the training process, a strategy supported by recent research ~\cite{fu2023deep, pan2025roboduet}. To ensure adaptability to different payloads, the end-effector's mass is also randomized during training, improving the policy's ability to handle objects of unknown weight. A detailed breakdown of all randomization parameters and the key reward components is provided in the Appendix~\ref{app: domain random}.

\subsection{Simulation Benchmark}
\label{sec:bench}

To evaluate navigation, manipulation, and whole-body control as a unified system, we present the first simulation benchmark tailored for long-horizon mobile manipulation in both indoor and outdoor environments.

\subsubsection{Asset and Scene Library}

To support realistic and versatile evaluation environments, we curate a diverse set of assets encompassing both object instances and full-scale 3D scenes. The object assets are sourced from a combination of prior open-source datasets~\cite{wang2024grutopia, 2025geniesim, nasiriany2024robocasa}, publicly available object repositories, and manually created models.

\textit{Object Assets}: We curate a diverse set of interactive objects categorized into four types: 50 rigid objects (e.g., common graspable items), 15 containers (e.g., bowls and bins with annotated containment region), 30 articulated structures (e.g., cabinets and doors), and 10 draggable items (e.g., carts and chairs). 

\textit{Environments}: Our benchmark includes 10 realistic scenes, with 5 indoor homes, 2 supermarkets, 1 restaurant, and 2 outdoor courtyards featuring slopes and stairs. All environments are designed for full traversability by legged robots and support multiple initialization zones to allow sampling and spatial variation of large-scale tasks.

\subsubsection{Rich Domain-style Variation}

To ensure generalization, we incorporate variability across four dimensions during simulation rollout: (1) \textit{Object layouts} are varied within semantic constraints across episodes, promoting diversity in interaction. (2) \textit{Physical attributes}, including mass, friction, and articulation limits, are resampled per episode to induce dynamic variability. (3) \textit{Environment conditions} such as lighting, material textures, and clutter elements are randomized to simulate perceptual noise. (4) \textit{Terrain complexity} is varied across outdoor scenes to assess locomotion robustness.

\subsubsection{Multi-stage Task Suite}

Our benchmark includes two categories of tasks: short-horizon manipulation skills merged from ARNOLD~\cite{gong2023arnold}, and long-horizon mobile manipulation tasks designed to reflect practical daily scenarios.

\textit{Short-Horizon Arnold Tasks.} We integrate four single-step manipulation tasks from the ARNOLD benchmark: \textsc{PickupObject}, \textsc{ReorientObject}, \textsc{OpenCabinet}, and \textsc{CloseCabinet}. While retaining their original goal state definitions and scene configurations, we adjust the spatial layout and object positioning to accommodate the kinematics and workspace of our quadruped robot platform, ensuring fair and consistent evaluation.

\textit{Long-Horizon Mobile Manipulation.}  
To assess the system's embodied reasoning, navigation, and sequential manipulation capabilities, we construct 8 multi-stage tasks spanning diverse indoor and outdoor scenarios. Each task consists of 2–3 subgoals, with a total of 246 indoor and 58 outdoor variations spanning object types, spatial layouts, and interaction modes.

Our task pool emphasizes a broad range of skills spanning grasping, reorientation, container placement, articulated manipulation, and long-term navigation over complex terrain. The combination of short and long-horizon tasks enables benchmarking at both low-level manipulation and high-level planning. Detailed task configurations are elaborated in Appendix~\ref{app: benchmark}.

\subsubsection{Modular Evaluation Protocol}

We evaluate both overall task success and per-action success rate. For instance, in the \textsc{CartDelivery} task, we define subtasks such as \textit{nav\_to\_object}, \textit{pick\_object}, \textit{nav\_to\_cart}, \textit{place\_object}, \textit{drag\_cart}, and \textit{nav\_to\_goal}. We determine actions success by monitoring the robot and cart’s world poses, as well as the relative pose between the object and the cart. A subtask is considered complete if its corresponding goal condition is met during the task horizon. This protocol captures both execution precision and planning consistency.

\section{Experiment}
\subsection{High-level Planner Performance}
\label{sec: exp_planner}

To evaluate the performance of our high-level planner in a modular and scalable manner, we conducted experiments based on the benchmark discussed in Section~\ref{sec:bench}. Firstly, we tested our local planner on thousands of single-step test cases, focusing on precision and consistency. Secondly, we integrated the global planner and evaluated our proposed approach on several hundred long-horizon mobile manipulation tasks. Additionally, we performed a detailed analysis of the completion rates for the decomposed atomic actions within each task.

\subsubsection{ARNOLD Short-horizon Tasks}

Before moving on to the long-horizon evaluation, we first conducted experiments in relatively confined spaces to demonstrate the fine-grained operation precision and generalization capabilities of our framework. We migrated four short-horizon tasks from ARNOLD and faithfully replicated their continuous monitoring system for goal states.

Their evaluation protocol divides five splits into two categories: 1) \textit{Seen} includes shuffled seen data; 2) \textit{Novel} features one of unseen components (objects, scenes, or goal states). We compare against their strongest baseline model PerAct~\cite{shridhar2022perceiveractormultitasktransformerrobotic}, an end-to-end imitation learning paradigm trained on large-scale human trajectories, which leverages observations from five external cameras to achieve accurate spatial perception. As shown in Table \ref{tab:short-horizon performance}, our method achieves substantial overall improvements, demonstrating superior fine-grained manipulation capabilities while relying solely on a single egocentric camera. Moreover, while their performance declined dramatically on novel splits, our method maintains stable performance across all datasets, showcasing generalized ability to handle O.O.D object configurations. Further experiment details are provided in Appendix~\ref{app: exp_short}.

\begin{table}[htbp]
\centering
\begin{tabular}{
    >{\centering\arraybackslash}p{1.4cm}
    >{\centering\arraybackslash}p{1.2cm}
    >{\centering\arraybackslash}p{1.2cm}
    >{\centering\arraybackslash}p{1.2cm}
    >{\centering\arraybackslash}p{1.2cm}
}
\toprule
\multirow{2}{*}{} & \multicolumn{2}{c}{\textbf{Seen}} & \multicolumn{2}{c}{\textbf{Novel}} \\
                  & PerAct & Ours & PerAct & Ours \\
\midrule
\textsc{P.Object}   & 94.03 & 60.45 & 25.70 & \textbf{45.24} \\
\textsc{R.Object}   & 19.48 & \textbf{51.32} & 8.23 & \textbf{52.09} \\
\textsc{O.Cabinet}  & 31.09 & \textbf{56.30} & 16.62 & \textbf{51.09} \\
\textsc{C.Cabinet}  & 60.81 & \textbf{74.32} & 41.32 & \textbf{79.50} \\
\bottomrule
\end{tabular}
\caption{Performance comparison between our approach and PerAct on 4 ARNOLD tasks, evaluated on both seen and generalized splits by success rate (\%).}
\label{tab:short-horizon performance}
\end{table}

\subsubsection{ODYSSEY Long-horizon Tasks}
\begin{table*}[h]
\centering
\begin{tabular}{@{}l|*{6}{>{\centering\arraybackslash}p{1.6cm}}|*{5}{>{\centering\arraybackslash}p{1.8cm}}@{}}
\toprule
\textbf{} & \textsc{I.Collect} & \textsc{R.Navigate} & \textsc{C.Delivery} & \textsc{C.Storage} & \textsc{Restocking} & \textsc{Shopping} & \textsc{O.Collect} & \textsc{O.Delivery} \\
\midrule
Navigate & 97.4 & 86.6 & 98.3 & 97.7 & 98.2 & 98.3 & 98.4 & 95.6\\
Pick & 72.7 & / & 84.6 & 79.6 & 83.3 & 85.0 & 69.0 & 72.7\\
Place & 96.8 & / & 72.7 & 83.8 & 79.2 & 76.5 & 95.0 & 80.0\\
Push/Pull & / & 94.1 & / & 71.0 & / & / & / & 85.7\\
Drag & / & / & 69.2 & / & / & 79.2 & / & /\\
\textbf{Overall} & \textbf{66.7} & \textbf{69.8} & \textbf{41.0} & \textbf{44.9} & \textbf{56.7} & \textbf{47.5} & \textbf{63.3} & \textbf{46.4}\\
\bottomrule
\end{tabular}
\caption{Overall success rates (\%) of 8 ODYSSEY long-horizon tasks, along with per-action success rates for each task.}
\label{tab: long-horizon performance}
\end{table*}

Table~\ref{tab: long-horizon performance} summarizes the performance of our system across eight long-horizon mobile manipulation tasks, reporting both overall task success rates and success rates for decomposed atomic actions. Notably, ODYSSEY consistently achieves 40\% or higher overall success across all tasks, and maintains over 60\% success in each atomic skill category, demonstrating robust coordination in a generalized long-horizon task. Building on this, we highlight several key findings from different perspectives of system performance:

\textbf{Low-level ability:} The consistent success rates across indoor and outdoor settings, despite the presence of irregular terrains, validate the reliable locomotion and effective pose tracking enabled by our terrain-adaptive whole-body control policy. Most control-related failures are caused by interactions with objects positioned beyond the robot's reachable range.

\textbf{Fine-grained action:} VLM-guided grounding enables high \texttt{pick} and \texttt{place} success rates across all tasks, demonstrating strong capability in identifying and localizing semantic targets. A large portion of failures stem from suboptimal gripper alignment, indicating limitations in the model’s spatial reasoning over object geometry. In addition, tasks involving more complex interactions such as \texttt{drag} and \texttt{pull} occasionally fail due to inaccurate localization, particularly with slim handles or partially occluded items.

\textbf{Task-level planning:} Our global task planner demonstrates strong symbolic reasoning over instance graphs, enabling reliable multi-stage task decomposition. In conjunction, our SLAM-based path planner ensures safe and consistent navigation. These components together lead to high \texttt{navigate} success rates across all tasks.

\subsection{Low-level Policy Performance}
We evaluated the proposed whole-body control policy against RoboDuet~\cite{pan2025roboduet}, a baseline that also uses a two-stage training process.  In contrast to RoboDuet's dual-policy (locomotion and manipulation) approach with base-centric sampling, our method employs a single, unified policy trained with a novel terrain-invariant end-effector sampling strategy. For the evaluation, 4096 parallel agents were instantiated with five data samples collected from each agent. 

\subsubsection{Metric} 
To quantitatively evaluate performance in the simulator, the following metrics are defined:

\begin{itemize}
    \item \textbf{Base Tracking Error}: The error between commanded and actual base velocities, comprising linear ($e_{x}$, $e_{y}$) and angular $e_\omega$ components.
    \item \textbf{End-Effector Position Error}: The Euclidean distance ($D_{pos}$) between the current and commanded end-effector positions in the world frame.
    \item \textbf{End-Effector Orientation Error}: The quaternion geodesic distance ($D_{ori}$) between the current $\mathbf{q}_{curr}$ and target $\mathbf{q}_{tar}$ orientations, calculated as $D_{ori} = 2*arcos(|\mathbf{q}_{curr} \cdot \mathbf{q}_{tar}|)$.
    
\end{itemize}

\begin{table}[h]
\centering
\begin{tabular}{@{}ccccc@{}}
\toprule
   & \multicolumn{2}{c}{\textbf{Stand still}} & \multicolumn{2}{c}{\textbf{Move}} \\ \midrule
   & Roboduet        & ours          & Roboduet & Ours          \\
$e_{x}\downarrow$  & 0.32            & \textbf{0.08}          & 9.70      & \textbf{0.36} \\
$e_{y}\downarrow$ & \textbf{0.34}            & 2.69          & 15.42    & \textbf{2.31} \\
$e_{\omega}\downarrow$  & 0.32            & \textbf{0.26}         & 60.59    & \textbf{0.79} \\
$D_{pos}\downarrow$ & \textbf{11.08}           & 11.48        & 10.75    & \textbf{10.57} \\
$D_{ori}\downarrow$ & 47.14           & \textbf{46.93}        & 47.53    & \textbf{47.15} \\ \bottomrule
\end{tabular}
\caption{The quantitative result under static (stand still) and dynamic (move) conditions.}
\label{tab: policy sim eval}
\end{table}

\begin{table}[h]
\centering
\begin{tabular}{@{}cccc@{}}
\toprule
      & \multicolumn{2}{c}{\textbf{Train}}                     & \textbf{Evaluation}            \\ \midrule
      & RoboDuet              & Ours                  &                       \\ 
$\hat{x}$    & {[}-1.00, 1.00{]}     & {[}-1.00, 1.00{]}     & {[}-1.50, 1.50{]}     \\
$\hat{y}$    & {[}0.00, 0.00{]}      & {[}-1.00, 1.00{]}     & {[}0.00, 0.00{]}      \\
$\hat{w}$   & {[}-0.60, 0.60{]}     & {[}-1.00, 1.00{]}     & {[}-1.50, 1.50{]}     \\
$\hat{l}_{ee}$    & {[}0.30, 0.70{]}      & {[}0.30, 0.65{]}      & {[}0.20,0.80{]}         \\
$\hat{p}_{ee}$     & {[}-0.45$\pi$, 0.45$\pi${]} & {[}-0.17$\pi$, 0.33$\pi${]} & {[}-0.50$\pi$, 0.50$\pi${]} \\
$\hat{y}_{ee}$    & {[}-0.50$\pi$, 0.50$\pi${]} & {[}-0.33$\pi$, 0.33$\pi${]} & {[}-0.50$\pi$, 0.50$\pi${]} \\
$\hat{\alpha}_{ee}$ & {[}-0.45$\pi$, 0.45$\pi${]} & {[}-0.50$\pi$, 0.50$\pi${]} & {[}-0.50$\pi$, 0.50$\pi${]} \\
$\hat{\beta}_{ee}$  & {[}-0.33$\pi$, 0.33$\pi${]} & {[}-0.17$\pi$, 0.50$\pi${]} & {[}-0.50$\pi$, 0.50$\pi${]} \\
$\hat{\gamma}_{ee}$  & {[}-0.42$\pi$, 0.42$\pi${]} & {[}-0.50$\pi$, 0.50$\pi${]} & {[}-0.50$\pi$, 0.50$\pi${]} \\ \bottomrule
\end{tabular}
\caption{Sampling ranges for commands used during training and evaluation. The locomotion commands consist of linear velocities ($\hat{x}$, $\hat{y}$) and angular velocity ($\hat{w}$). The end-effector target is defined by a position in spherical coordinates (radius $\hat{l}_{ee}$, pitch $\hat{p}_{ee}$ and yaw $\hat{y}_{ee}$) and an orientation in Euler angles ($\hat{\alpha}_{ee}$, $\hat{\beta}_{ee}$, $\hat{\gamma}_{ee}$). }
\label{tab: ee sample}
\end{table}

\subsubsection{Simulation result}
Our method was compared with evaluations conducted under both static (standing) and dynamic (moving) conditions. Due to the robot's structural constraints, sampling unreachable samples leads to self-collisions, which negatively impact the training process. To mitigate this issue, we deliberately reduced the volume of the sampling space used during training. To ensure a fair comparison and test generalization, both methods were evaluated in the same, larger workspace, as detailed in Table \ref{tab: ee sample}, with the comparative results of this evaluation presented in Table \ref{tab: policy sim eval}.

The evaluation results indicate that our policy achieves better performance in base velocity tracking (rows 1-3), an improvement we attribute to the inclusion of terrain data in the policy's observation, which enhances the robot's state estimation. End-effector pose tracking performance remains comparable to the baseline (rows 4-5). Notably, a key aspect of this evaluation is that our policy was trained in an end-effector workspace intentionally smaller than that used for the baseline, and our policy is adaptive to different topographies (e.g., steps). This demonstrates strong generalization capabilities from a more constrained training domain of our approach.

\subsection{Sim-to-real Performance}
We conducted real-world experiments to validate the sim-to-real performance of our framework, which integrates the high-level coarse-to-fine task planner with the low-level whole-body control policy.

\subsubsection{Robot System Setup}
Our robot platform, as shown in Fig.~\ref{fig: sim2real}, combines a 12-DoF Unitree Go2 quadruped with a 6-DoF Arx5 manipulator. The Go2 (15kg weight, 8kg payload) includes a built-in Unitree L1 LiDAR, and the 3.35kg Arx5 arm is mounted on its back, similar to ~\cite{ha2024umi}. For high-level perception, the platform is equipped with a MID-360 LiDAR for localization and two RealSense cameras: a head-mounted D435i for RGB imagery and a gripper-mounted D405 for RGB-D data. The control policy operates at 50 Hz, with a PD controller issuing motor commands at 200 Hz.


\subsubsection{Real-world experiments}
\begin{figure}[H]
    \centering
    \includegraphics[width=1\linewidth]{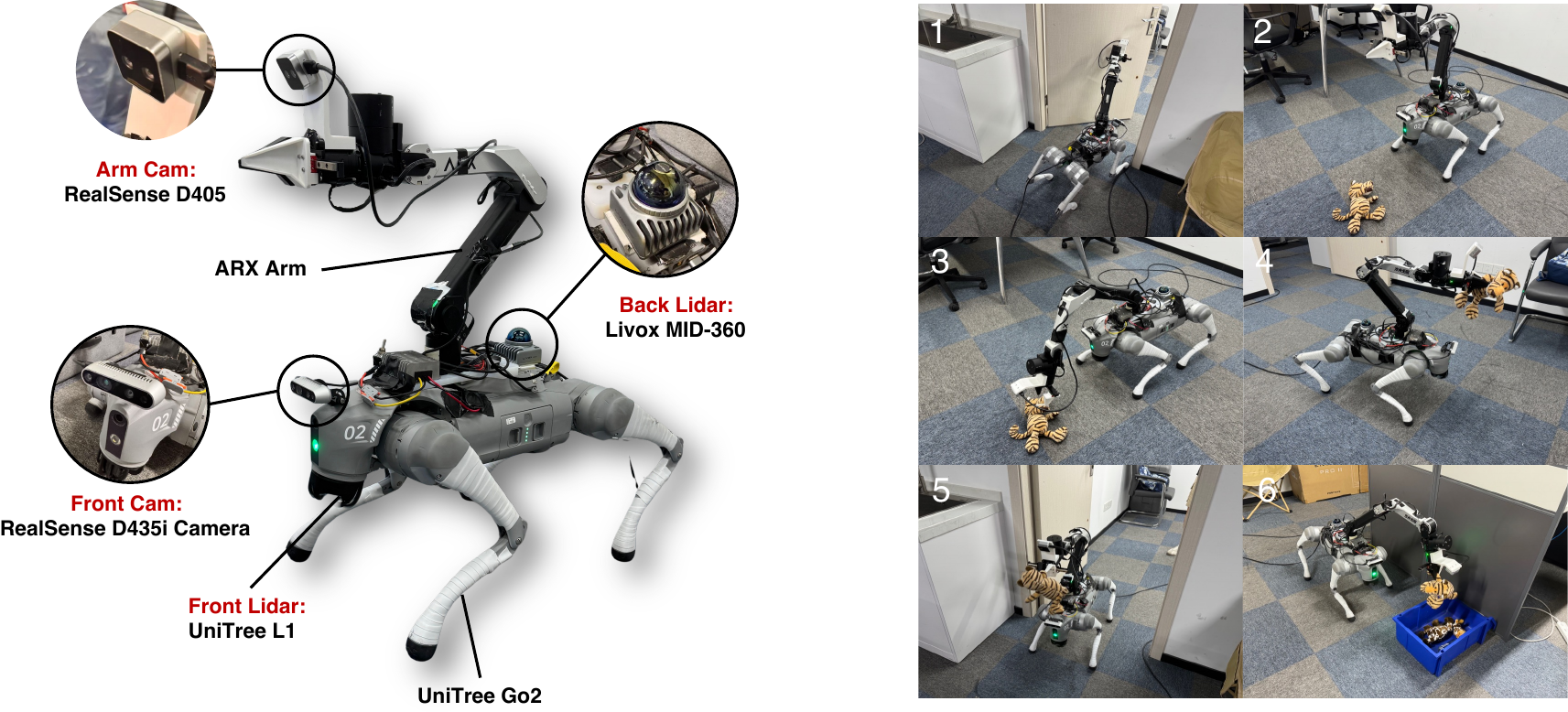} 
    \caption{The robot system and real-world experiments}
    \label{fig: sim2real}
\end{figure}

The ODYSSEY framework was evaluated on two long-term tasks ("navigate to pick" and "pick and place") using five different objects. The entire system demonstrated successful sim-to-real transfer on task planning and execution as illusion in Fig.~\ref{fig: sim2real}.

Despite this success, some sim-to-real gaps persist. For instance, the robot occasionally failed at grasping small objects due to the inaccuracies in end-effector tracking and visual perception. Through these experiments, our approach has demonstrated significant potential for solving long-horizon mobile exploration and manipulation tasks, while simultaneously identifying the primary challenges—robust perception and high-precision control—that must be addressed for seamless real-world deployment.

\section{Conclusions and Future Work}
We present ODYSSEY, a unified framework for open-world mobile manipulation that integrates hierarchical task planning with terrain-adaptive whole-body control. Our approach demonstrates robust sim-to-real transfer and generalization across diverse environments and long-horizon tasks. Future work will extend our benchmark into a comprehensive evaluation paradigm for vision-language models (VLMs) and mobile manipulators, enabling cross-embodiment assessment of semantic reasoning and locomotion-manipulation coordination. Additionally, we aim to explore the emergent capabilities of active perception, where dynamic scene understanding and adaptive motion synergize for more efficient real-world interaction. This direction could unlock new behaviors in cluttered, unstructured environments, further bridging the gap between high-level planning and low-level control.

\newpage
\bibliography{aaai2026}



\clearpage
\renewcommand{\thesection}{\Alph{section}}  
\appendix
\section*{Appendix}

\section {Long-horizon Planner Pipeline}
\subsection{Instance-Level Semantic Graph Construction}
\label{app: graph constructor}

\begin{figure}[h]
  \centering
  \includegraphics[width=1.0\linewidth]{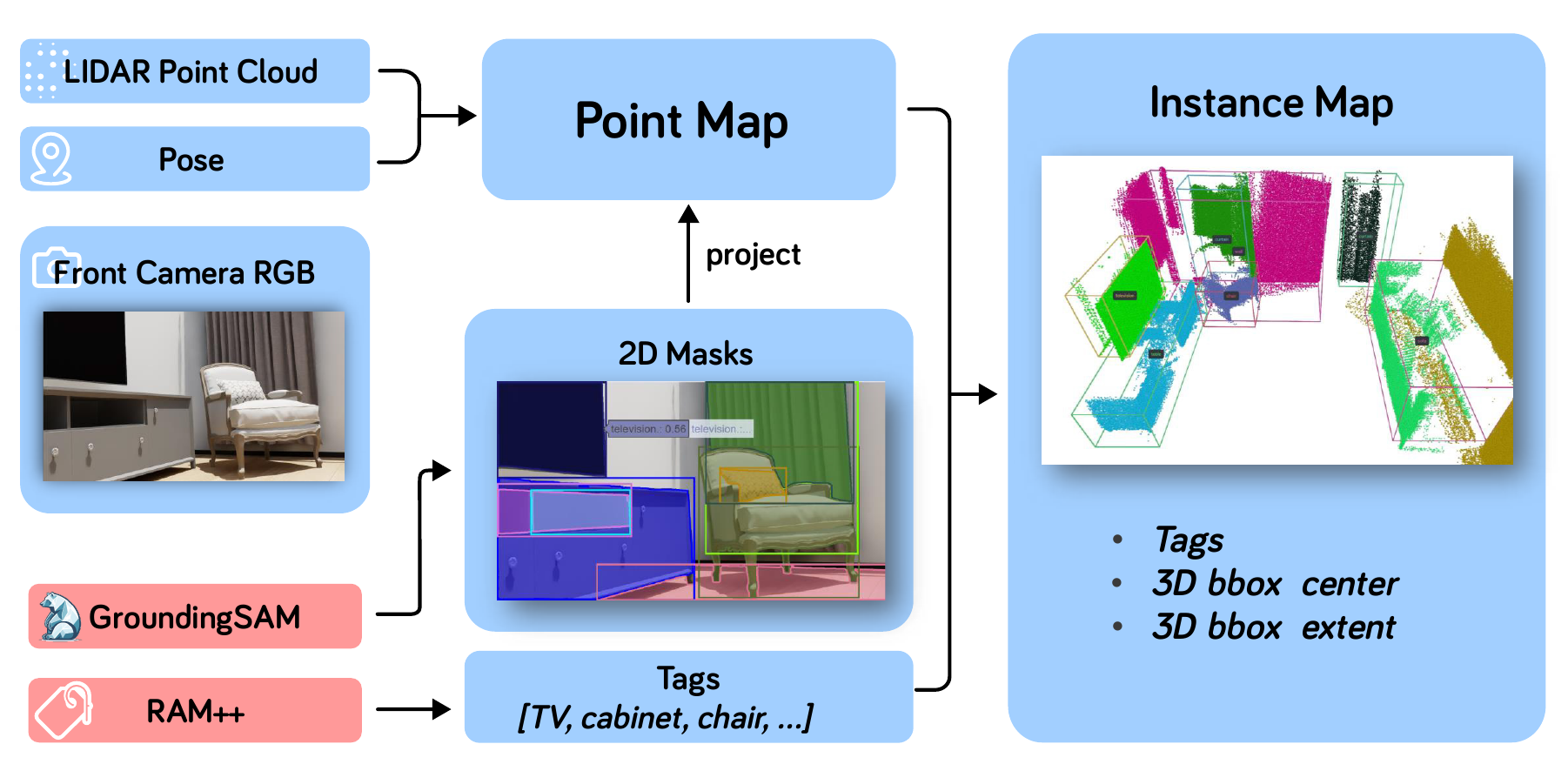}
  \caption{Instance-level graph construct pipeline.}
  \label{fig:llm_prompt_examples}
\end{figure}

To enable egocentric task planning, we implement an online multi-modal perception pipeline that constructs an instance-level semantic graph from onboard sensors.

A downward-facing LiDAR provides real-time point clouds at 10\,Hz, which are aggregated into a global point map using the robot's estimated pose. This map serves both as a geometric substrate for semantic abstraction and as a SLAM-based 2D occupancy grid for path planning.

In parallel, a front-facing RGB camera captures visual frames. We use RAM++~\cite{huang2023open} to identify candidate object labels from each RGB image. These labels, along with the image, are fed into Grounding-SAM~\cite{ren2024grounded} to obtain instance-level segmentation masks $m_i$. We replace the standard SAM model with MobileSAM\cite{zhang2023mobilesamv2fastersegment} to improve efficiency. The masks are projected onto the current LiDAR frame to obtain per-object point cloud segments.

Inspired by prior works\cite{gu2024conceptgraphs, jatavallabhula2023conceptfusion}, to enable consistent object tracking across time, we fuse observations using both semantic and geometric similarity:

\begin{itemize}
    \item \textbf{}{Semantic Similarity.} For each instance mask, we extract a sparse visual-semantic feature map $f_I$ using MaskCLIP~\cite{zhou2022extract}. We pool features within each mask $m_i$ to generate an instance-level descriptor $\mathbf{f}_i \in \mathbb{R}^d$. Two instances $i$ and $j$ are considered semantically consistent if their cosine similarity satisfies:
    \begin{equation}
    \text{sim}_{\text{sem}}(\mathbf{f}_i, \mathbf{f}_j) = \frac{\mathbf{f}_i^\top \mathbf{f}_j}{\|\mathbf{f}_i\|\|\mathbf{f}_j\|} > \tau_\text{sem}, \quad \tau_\text{sem} = 0.8.
    \end{equation}
    
    \item \textbf{}{Geometric Similarity.}
    Let $\mathcal{P}_i$ and $\mathcal{P}_j$ be the point sets of two object candidates. We consider them geometrically aligned if at least 80\% of the points in $\mathcal{P}_i$ have a nearest neighbor in $\mathcal{P}_j$ within radius $\epsilon$:
    \begin{equation}
    \frac{1}{|\mathcal{P}_i|} \sum_{\mathbf{p} \in \mathcal{P}_i} \mathbb{I} \left[ \min_{\mathbf{q} \in \mathcal{P}_j} \|\mathbf{p} - \mathbf{q}\| < \epsilon \right] > \tau_\text{geo}, \quad \tau_\text{geo} = 0.8.
    \end{equation}
\end{itemize}

When both criteria are satisfied, the two instances are merged as the same object across time.

The final instance graph $G = \{O_i\}$ contains nodes $O_i$ with attributes including semantic tags and 3D bounding boxes. This symbolic representation serves as the input for downstream task decomposition and decision-making.

\subsection{Hierarchical Planning Prompts}
\subsubsection{LLM-based Task-level Prompts}
This section illustrates how we leverage large language models (LLMs) to decompose tasks into structured symbolic action sequences (e.g., \texttt{navigate}, \texttt{pick}, \texttt{place}). The prompt input consists of the user instruction and a summarized semantic scene graph. Example prompts are shown in Fig.~\ref{fig:llm_prompt_examples}.

\subsubsection{VLM-based Manipulation Prompts}
\begin{figure*}[h]
  \centering
  \includegraphics[width=1.0\linewidth]{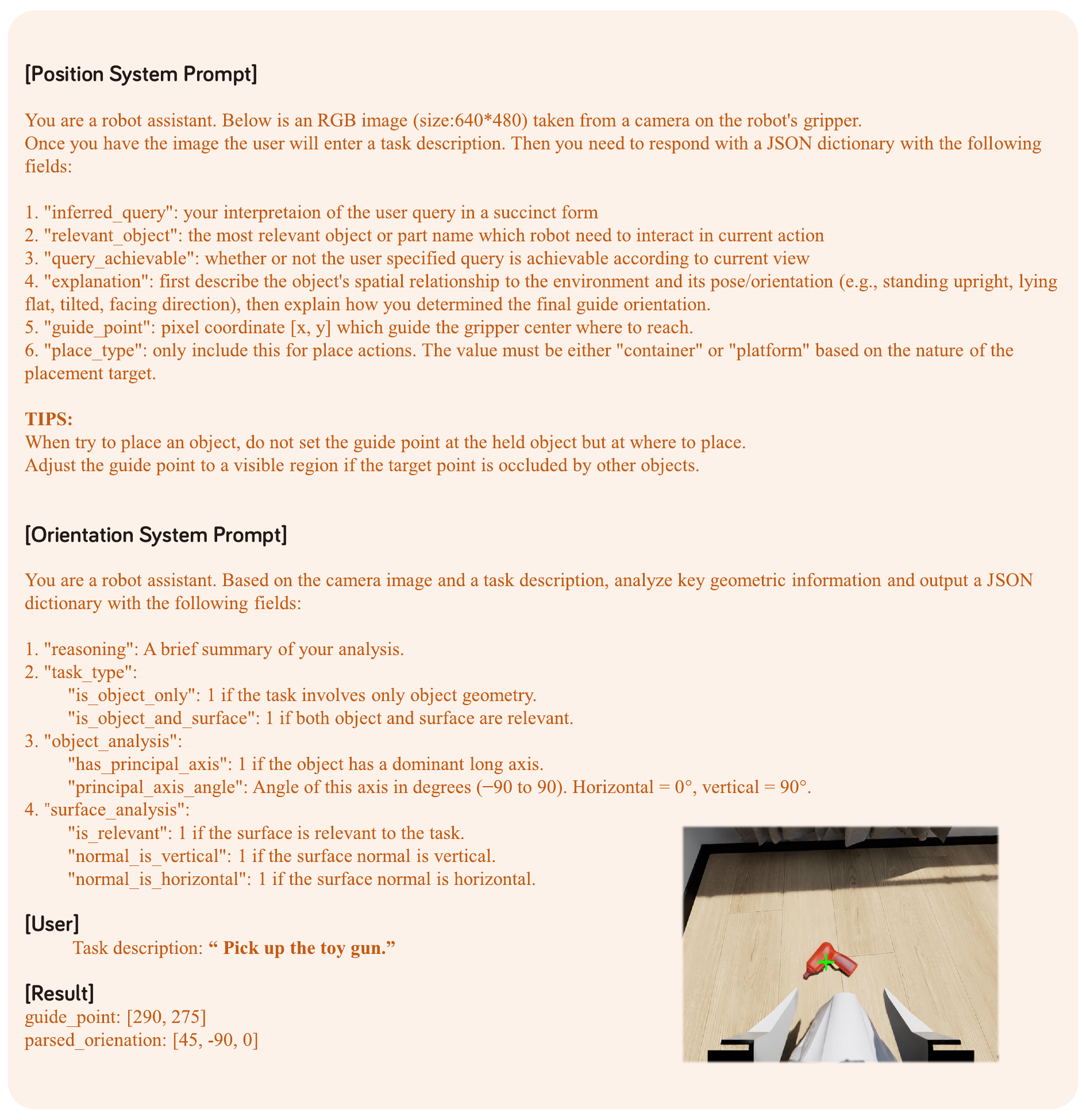}
  \caption{Prompt examples for local manipulation guidance using VLM. Given an egocentric image and sub-task description, the model predicts a contact point and end-effector orientation under geometric constraints.}
  \label{fig:vlm_prompt_examples}
\end{figure*}

For the execution of atomic manipulation actions, we demonstrate how vision-language models (VLMs) are used to infer task-relevant contact points and end-effector orientations. The prompt input includes a first-person RGB observation and a natural language description of the current sub-task. Examples are shown in Fig.~\ref{fig:vlm_prompt_examples}.

\section {Whole-body Policy Training}
\subsection{Reward design} 
\label{app: reward design}

This section details the key reward terms used during policy training, summarized in Table \ref{app tab: reward design}. 

\begin{table}[H]
\centering
\begin{tabular}{@{}cccc@{}}
\toprule
              &          & \multicolumn{2}{c}{\textbf{Weight}} \\ 
\textbf{Term}          & \textbf{Equation} & \textbf{Stage 1}      & \textbf{Stage 2}     \\ \midrule
$r^{track}_{xy}$  & $\exp(-((\hat{\mathbf{v}}_{xy} - \mathbf{v}_{xy}))/\gamma_{xy})$        & 2.75            & 2.75           \\
$r^{track}_{yaw}$  & $\exp(-((\hat{\omega} - \omega))/\gamma_{\omega})$        & 1.50            & 1.50           \\
$r^{track}_{ee-pos}$  & $\sqrt{(\hat{\mathbf{p}}_{ee} - \mathbf{p}_{ee})^2}$        & 0.00            & -1.20           \\
$r^{track}_{ee-ori}$  & $\sqrt{(\hat{\mathbf{\phi}}_{ee} - \mathbf{\phi}_{ee})^2}$        & 0.00            & -1.50           \\
$r_{gait}$   & Eq. \ref{app eq: gait rew}        & 0.75            & 0.75           \\
$r_{freq}$    & Eq. \ref{app eq: freq rew}        & 1.25e1            & 1.25e1           \\ \midrule

$r_{torque}^{base}$  & $|\mathbf{\tau}^{base}|^2$        & -2.0e-4            & -2.0e-4           \\
$r_{acc}^{base}$    & $|\mathbf{\ddot{q}}^{base}|^2$        & -2.5e-7            & -2.0e-7          \\
$r_{power}^{base}$  & $ |\mathbf{\tau}^{base}| * |\mathbf{\dot{\mathbf{q}}}^{base}|$        & -2.0e-5            & -2.0e-5           \\
$r_{torque}^{arm}$  & $|\mathbf{\tau}^{arm}|^2$        & 0.00            & -4.0e-4             \\
$r_{acc}^{arm}$    & $|\mathbf{\ddot{q}}^{arm}|^2$        & 0.00            & -2.5e-6           \\
$r_{power}^{arm}$  & $ |\mathbf{\tau}^{arm}| * |\mathbf{\dot{\mathbf{q}}}^{arm}|$        & 0.00            & -2.0e-4           \\

$r_{smooth}$ & $\sqrt{(\mathbf{a}_t -\mathbf{a}_{t-1})^2}$        & -0.02            & -0.02           \\ \bottomrule
\end{tabular}
\caption{The key reward terms used in policy training}
\label{app tab: reward design}
\end{table}

To encourage a stable gait, we use a gait reward $r_{gait}$ that promotes specific foot contact patterns. It encourages synchrony between diagonal leg pairs (FL/RR, FR/RL) using a reward term $r_s$ and asynchrony between all other pairs using $r_a$. The final gait reward is a multiplicative combination of individual reward terms ($r_s, r_a$), which are functions of the squared error between the time two legs (A, B) have spent in the air ($A_{air}$) or in contact ($A_{cont}$).
\begin{align}
    t_{air}^{s}(A, B) &= clip( (A_{air} - B_{air})^2, 0.0, 0.04)\\
    t_{cont}^{s}(A, B) &= clip( (A_{cont} - B_{cont})^2, 0.0, 0.04) \\
    r_s(FL, RR) &= e^{-( t_{air}^{s}(FL, RR) + t_{cont}^{s}(FL, RR) )} \\
    t_{air}^{a}(A, B) &= clip((A_{air} - B_{cont})^2, 0.0, 0.04) \\ 
    t_{cont}^{a}(A, B) &=clip((A_{cont} - B_{air})^2, 0.0, 0.04) \\ 
    r_a(FL, FR) &= e^{-( t_{air}^{a}(FL, FR) + t_{cont}^{a}(FL, FR) )}  \\
    r_{gait} &= \prod_{i,j \in \text{sync pairs}} r_{s}(i, j) \cdot \prod_{k,l \in \text{async pairs}} r_{a}(k, l) \label{app eq: gait rew}
\end{align} 
A frequency reward $r_{fre}$ regulates the gait's cadence by penalizing deviation from a target frequency $f_{target}$. The frequency for each leg $f(leg)$ is the inverse of the time between consecutive ground contacts ($t_k^{cont} - t_{k-1}^{cont}$), where $k$ is the timestamp. The frequency reward $r_{fre}$ is defined as bellow:
\begin{align}
    r_{f}(\text{leg}) &= \exp(-0.5 \cdot (f(\text{leg}) - f_{target})^2) \\
    r_{fre} &= \prod_{\text{leg} \in {\text{FL,FR,RL,RR}}} r_f(\text{leg}) \label{app eq: freq rew}
\end{align}
In stage one, training focuses on locomotion. The reward function includes task-specific terms for base velocity tracking ($r_{xy}^{track}, r_{yaw}^{track}$) and gait shaping ($r_{gait}, r_{freq}$). To improve stability and facilitate sim-to-real transfer, several regularization rewards are applied. These include penalties on torque ($r_{torque}$), joint acceleration ($r_{acc}$) and action-to-action changes ($r_{smooth}$) to encourage stable policy outputs. Additionally, a power consumption penalty ($r_{power}$) is used to discourage excessive mechanical work. In stage two, all Stage one rewards are retained, and new task-specific terms are introduced for end-effector pose tracking (position $r_{ee-pos}^{track}$ and orientation $r_{ee_ori}^{track}$). Furthermore, the regularization penalties(torque, acceleration, and power) are extended to include the manipulator's joints.

\subsection{Domain randomization} 
\label{app: domain random}

\begin{table}[H]
\centering
\begin{tabular}{@{}ccc@{}}
\toprule
\textbf{Parameter}         & \textbf{Ranges} & \textbf{Method} \\ \midrule
Friction          & [0.4, 2.0]      &  -       \\
Base Mass         & [-5.0, 5.0]      & add       \\
Base Pushing      & $x$: [-0.5, 0.5]      & interval       \\
                  & $y$: [-0.5, 0.5]      &         \\
Actuator Gains    & [0.8, 1.2]      & scale       \\
Ee Link Mass & [0.0, 0.2]      & add       \\ \midrule
Joint reset       &  [0.5, 1.5]      & scale       \\
Base reset        & $x$: [-0.5, 0.5], $y$: [-0.5, 0.5]      &  add     \\ 
                  & $\omega: [-\pi, \pi]$, $v_x$: [-0.5, 0.5]     &        \\ 
                  & $v_y$: [-0.5, 0.5], $v_z$: [-0.5, 0.5]     &        \\ 
                  & $\alpha$: [-0.5, 0.5], $\beta$: [-0.5, 0.5]     &        \\ 
                  & $\gamma$: [-0.5, 0.5]     &        \\  \bottomrule
\end{tabular}
\caption{The domain randomization ranges. The method means the method of randomization, where "add" means add to the original, "scale" means scale up the original, and "interval" means continue for a while.}
\label{app tab: domain random}
\end{table}

To enhance out-of-distribution generalization and facilitate sim-to-real transfer, our policy is trained with extensive domain randomization, as detailed in Table~\ref{app tab: domain random}. This process includes randomizing the robot's initial conditions, such as its base pose and joint configuration, at the start of each episode. Crucially, to ensure adaptability to tasks involving different payloads, the mass of the end-effector is also randomized.

\section{Benchmark Configurations}
\label{app: benchmark}

This section presents the task setup used in our benchmark evaluation, including both short-horizon and long-horizon mobile manipulation tasks.

\subsection{ARNOLD Short-Horizon Tasks}

To evaluate the fine-grained manipulation capabilities of our system, we migrate four short-horizon tasks from the ARNOLD benchmark into our custom simulation environment. These tasks are designed to assess step-wise visuomotor precision and generalization under diverse configurations.

\subsubsection{Simulation Integration.}
To enable evaluation within our own simulation platform, we convert the raw ARNOLD data into standardized YAML configuration files. The conversion pipeline involves:

\begin{enumerate}
    \item Parsing each \texttt{.npz} sample to extract task-specific parameters, including scene setup, object category, pose, and robot initial state.
    \item Transforming all coordinate frames from ARNOLD’s convention to that of our simulator.
    \item Automatically generating natural language task instructions and adjusting the robot’s initial pose to ensure reachability and feasibility.
    \item Grouping tasks by scene and saving them as per-scene YAML files containing layout, object initialization, evaluation metrics, and interaction constraints.
\end{enumerate}

\begin{figure}[h]
  \centering
  \includegraphics[width=1.0\linewidth]{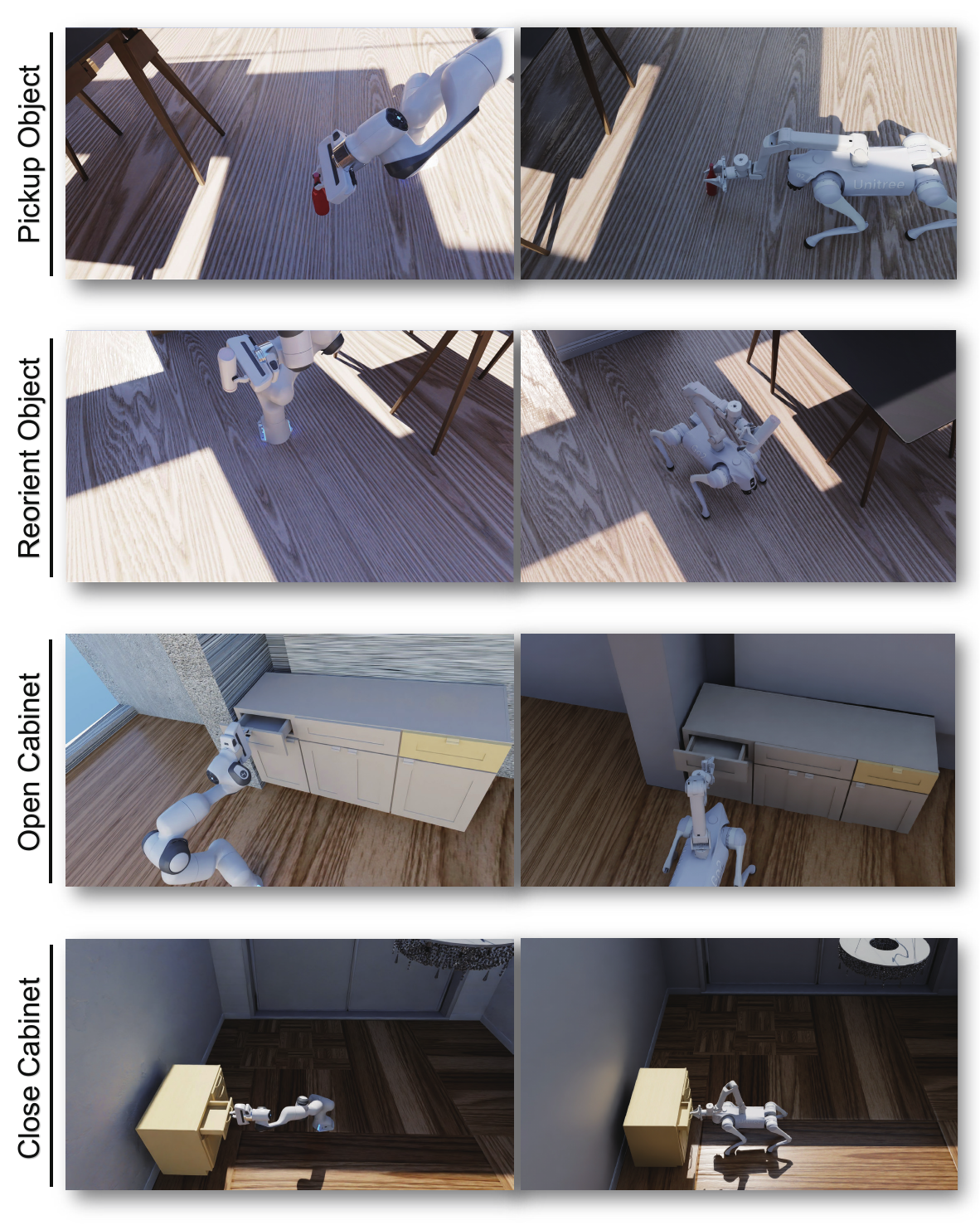}
  \caption{ARNOLD converted environments of four tasks.}
  \label{fig: arnold_convert}
\end{figure}

This setup ensures consistent large-scale deployment and reproducible benchmarking of ODYSSEY’s short-horizon manipulation performance across multiple scenarios. Examples of multiple tasks sharing a single scene configuration are shown in Figure~\ref{fig: arnold_convert}.

\subsubsection{Action Sequence Structure.}
To facilitate scalable benchmarking, we adopt predefined action sequences in place of LLM-based task decomposition. This ensures consistent execution patterns across trials and enables focused evaluation of local manipulation accuracy independent of high-level planning variance. For instance, the \textsc{OpenCabinet} task consists of the following four phases:

\begin{itemize}
    \item \textbf{Observe \& Plan:} Capture the scene (\textit{o1}) to identify the semantic target and compute the desired end-effector pose $a^*$ for contact.
    \item \textbf{Pre-Alignment:} Move the gripper to a pre-contact pose located 10\,cm away from $a^*$ along the approach axis (opposite to gripper's $z$-axis).
    \item \textbf{Execute Grasp:} Linearly advance the gripper along the approach direction to reach $a^*$ and close the fingers to secure the object.
    \item \textbf{Task Completion:} Apply the task-specific motion (e.g., pulling, lifting) until the environment state satisfies the success condition.
\end{itemize}

\begin{figure}[h]
  \centering
  \includegraphics[width=1.0\linewidth]{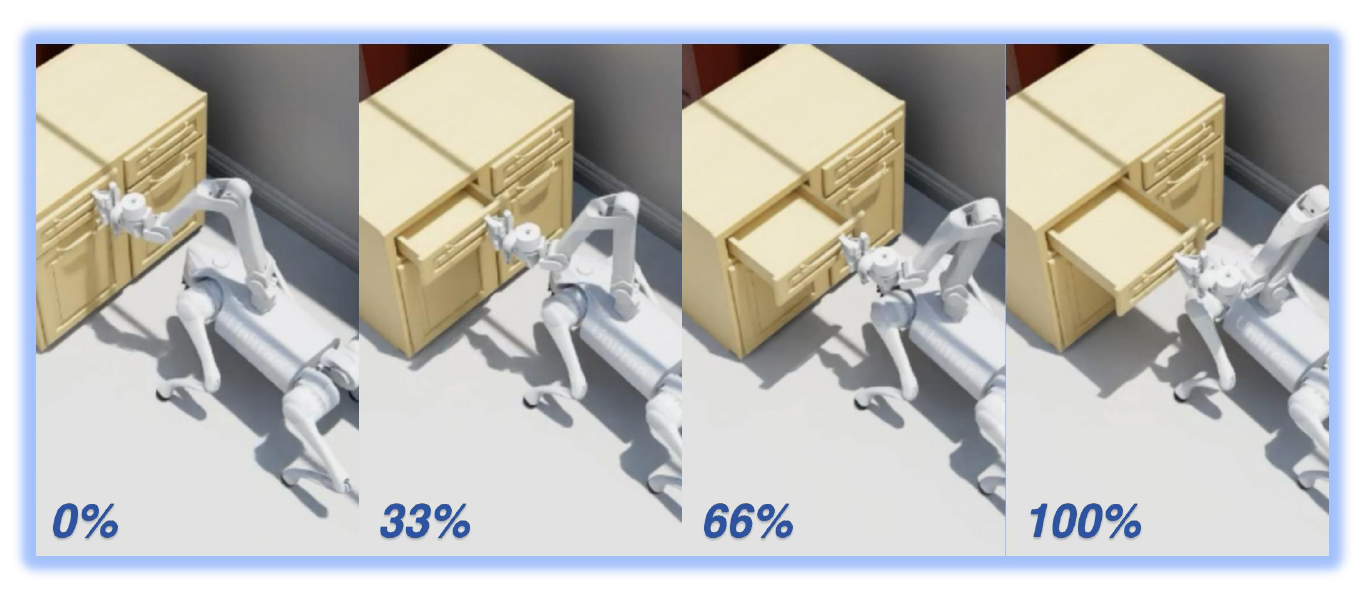}
  \caption{The continuous state monitoring of ARNOLD short-horizon tasks.}
  \label{fig: app_state_monitor}
\end{figure}

This structured formulation allows for fine-grained progress monitoring at 0\%, 33\%, 66\%, and 100\% completion stages, as illustrated in Figure~\ref{fig: app_state_monitor}.

\subsection{ODYSSEY Long-Horizon Tasks}
\begin{figure*}[h]
  \centering
  \includegraphics[width=0.95\linewidth]{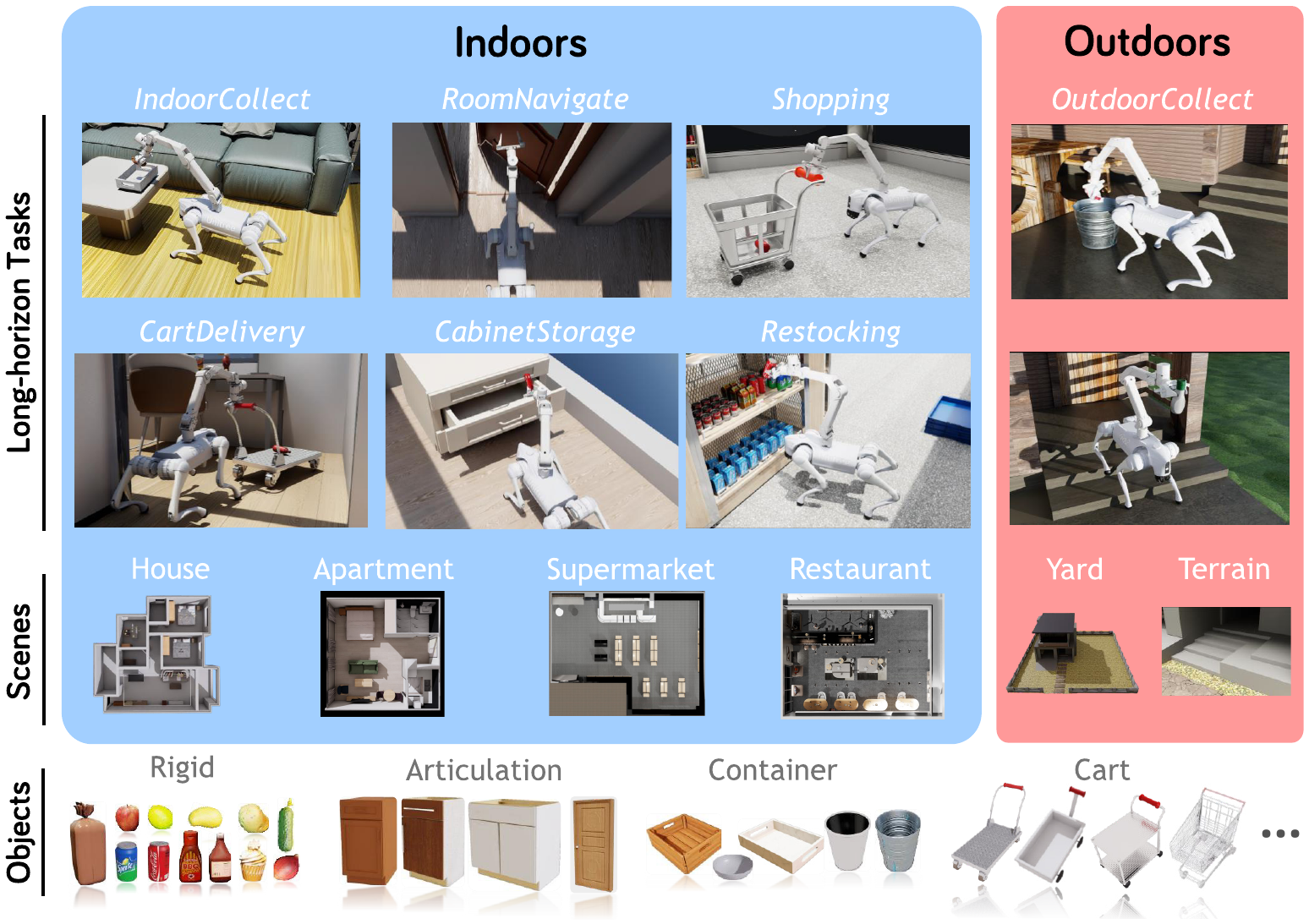}
  \caption{ODYSSEY benchmark details. Including task type, scene, and object configuration details.}
  \label{fig: app_benchmark}
\end{figure*}

As shown in Fig.~\ref{fig: app_benchmark}. Our benchmark includes eight long-horizon tasks constructed to evaluate full-pipeline performance, spanning perception, task planning, navigation, and manipulation. Each task consists of 2–3 subgoals and is designed with diverse object types, spatial distributions, and physical interaction modes.

Table~\ref{tab:long_task_config} summarizes the tasks, listing the task name, number of variations, and number of unique asset types involved in each.

\begin{table}[h]
\centering
\begin{tabular}{@{}lll@{}}
\toprule
\textbf{Task Names} & \textbf{Variations} & \textbf{Assets} \\
\midrule
\textsc{IndoorCollect} & 45 & 65 \\
\textsc{RoomNavigation} & 43 & 11 \\
\textsc{CartDelivery} & 39 & 47 \\
\textsc{CabinetStorage} & 49 & 69 \\
\textsc{Restocking} & 30 & 34 \\
\textsc{Shopping} & 40 & 45 \\
\cmidrule(r){1-3}
\textsc{OutdoorCollect} & 30 & 47 \\
\textsc{OutdoorDelivery} & 28 & 48 \\
\bottomrule
\end{tabular}
\caption{Overview of 8 long-horizon tasks with names, variation counts, and asset usage per task.}
\label{tab:long_task_config}
\end{table}



\newpage
\section{Experiment Details}

\subsection{Short-Horizon Evaluation}
\label{app: exp_short}

\subsubsection{Data Split}
To assess generalization under various task variations, the ARNOLD benchmark defines five splits:
\begin{itemize}
    \item \textbf{Test}: All components—objects, scenes, and states—are seen during training.
    \item \textbf{Novel Object}: Involves unseen objects, with seen scenes and goal states.
    \item \textbf{Novel Scene}: Uses unseen environments while keeping objects and goals seen.
    \item \textbf{Novel State}: Includes novel goal configurations under seen objects and scenes.
    \item \textbf{Any State}: Targets fine-grained generalization to arbitrary continuous goals.
\end{itemize}

In Section~\ref{sec: exp_planner}, we report the average success rate over the last four splits as a unified \textbf{Novel} setting for simplicity and comparison clarity. Here in the appendix, we present detailed breakdowns across all five splits.

\subsubsection{Baselines and Evaluation Protocol}
In Section~\ref{sec: exp_planner}, we primarily compare ODYSSEY with PerAct, the strongest baseline from ARNOLD. For a more complete analysis, we extend the evaluation to include all three PerAct variants:
\begin{itemize}
    \item \textbf{PerAct}: The standard model based on voxelized RGB-D fusion and Perceiver Transformer.
    \item \textbf{PerAct$\dagger$}: Enhanced with value-function supervision to improve decision-making.
    \item \textbf{PerAct (MT)$\dagger$}: A multi-task trained variant leveraging more diverse training signals.
\end{itemize}

In addition, following the ARNOLD evaluation protocol, each model is evaluated in a two-phase setup:
\begin{itemize}
    \item \textbf{Phase I (End-to-End)}: Includes both grasping and subsequent manipulation.
    \item \textbf{Phase II (Manipulation Only)}: Uses ground-truth grasp poses to isolate manipulation performance.
\end{itemize}

This extended evaluation reveals the specific strengths and failure modes of ODYSSEY compared to baselines, particularly under generalization stress tests.

\subsubsection{Results and Failure Analysis}

Table~\ref{tab:detailed-short-horizon-tasks} presents detailed performance across all tasks and data splits. ODYSSEY consistently outperforms all PerAct variants, particularly in generalization settings involving novel objects, scenes, or goal states. The two-phase evaluation further highlights the robustness of our manipulation module when decoupled from grasping errors.

Common failure cases across tasks are largely attributed to (1) inaccurate semantic reasoning by the vision-language model (VLM), such as object orientation confusion or suboptimal contact point prediction, and (2) challenges in physical interaction execution, including occlusion, narrow geometry, and unstable force transfer.

\begin{itemize}
    \item \textsc{PickupObject:} Phase I failures mainly arise from misjudged object orientation or selecting hard-to-grasp regions (e.g., bottlenecks). These issues are largely resolved in phase II, validating the effectiveness of our pose-conditioned manipulation.

    \item \textsc{ReorientObject:} Errors stem from inaccurate initial pose estimation or weak grasps on thin sections, leading to slippage during reorientation. ODYSSEY exhibits robust generalization even under such dynamic conditions.

    \item \textsc{OpenCabinet:} Limited improvement in phase II suggests inherent difficulty in the manipulation itself. Failures often relate to misidentified handles, drawer confusion, or inaccurate grasp positioning.

    \item \textsc{CloseCabinet:} Performance is mainly affected by occlusion and poor camera framing. Handles are sometimes out of view during approach or are visually ambiguous due to lighting or textures.
\end{itemize}

In summary, ODYSSEY demonstrates strong semantic grounding, generalization, and precision control, particularly in high-complexity tasks. Remaining challenges highlight future directions in visual robustness and contact policy refinement.

\subsection{Long-Horizon Evaluation}
\label{app: exp_long}
\subsubsection{Failure Analysis}

To complement the long-horizon evaluation in Section~\ref{sec: exp_planner}, we present a failure mode analysis aggregated over all 6 indoor and 2 outdoor ODYSSEY tasks. Rather than reporting per-task breakdowns, we group failures by environment type and visualize them separately for indoor and outdoor scenarios.

Each failure is assigned to one of the following three categories:

\begin{itemize}
    \item \textbf{Reasoning Failure:} Errors arising from the LLM-based task planner or VLM-based local action advisor, including incorrect subgoal decomposition, ambiguous symbolic grounding, or mispredicted contact poses.
    \item \textbf{Control Failure:} Low-level execution issues related to either locomotion or manipulation, such as unstable whole-body control on uneven terrain or failed grasp execution due to poor alignment or force instability.
    \item \textbf{Navigation Failure:} Failures caused by inability to bring target objects into a feasible manipulation workspace, or collisions during path execution due to inaccurate waypoint planning or obstacle avoidance.
\end{itemize}

\begin{figure}[h]
    \centering
    \includegraphics[width=1.1\linewidth]{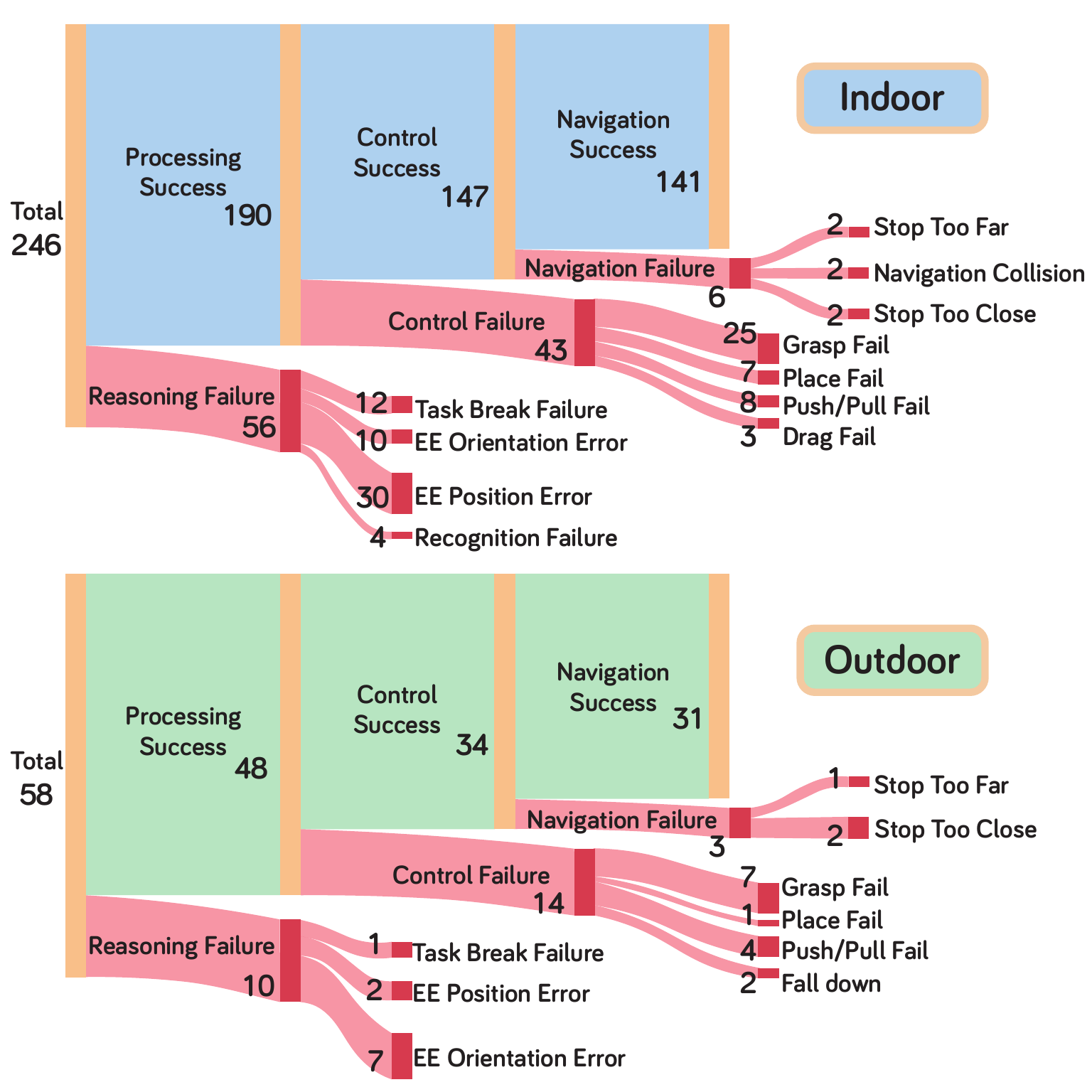}
    \caption{\textbf{Failure breakdown by environment type}. Failures are categorized into reasoning, control, and navigation errors. Indoor tasks suffer more from language and control failures, while navigation becomes more challenging in outdoor scenarios.}
    \label{fig:long_failure_breakdown}
\end{figure}

Figure~\ref{fig:long_failure_breakdown} shows the distribution of these failure types across indoor and outdoor tasks. From the analysis, we draw several key observations:

\begin{itemize}
    \item \textbf{Indoor tasks are dominated by reasoning and control failures}. Tasks such as \textsc{CabinetStorage} and \textsc{Restocking}, which involve complex semantic reasoning and precise manipulation, frequently fail due to incorrect subgoal decomposition or inaccurate contact prediction. Control-related issues, such as unstable grasps and force misalignment, are also prevalent due to tight spatial constraints and object diversity in indoor environments.
    
    \item \textbf{Control failures in outdoor scenarios are often tied to whole-body instability}. Unlike indoors, where control issues center on gripper-object interaction, outdoor control failures are more frequently caused by the robot falling or losing balance, particularly during manipulation actions performed on slopes or uneven surfaces.

    \item \textbf{Reasoning errors span both global and local levels}. Global planning failures include incorrect symbolic grounding or subgoal ordering, while local failures stem from misidentifying contact points or failing to detect objects due to occlusions or poor lighting.

\end{itemize}

\begin{table*}[h]
\centering
\begin{tabular}{
@{}l
>{\raggedright\arraybackslash}p{2.2cm}
>{\raggedleft\arraybackslash}p{1.2cm} >{\raggedleft\arraybackslash}p{1.2cm}
>{\raggedleft\arraybackslash}p{1.2cm} >{\raggedleft\arraybackslash}p{1.2cm}
>{\raggedleft\arraybackslash}p{1.2cm} >{\raggedleft\arraybackslash}p{1.2cm}
>{\raggedleft\arraybackslash}p{1.2cm} >{\raggedleft\arraybackslash}p{1.2cm}
@{}
}
\toprule
\textbf{Split} & \textbf{Method} &
\multicolumn{2}{c}{\textsc{PickupObject}} &
\multicolumn{2}{c}{\textsc{ReorientObject}} &
\multicolumn{2}{c}{\textsc{OpenCabinet}} &
\multicolumn{2}{c}{\textsc{CloseCabinet}} \\
\midrule
\multirow{4}{*}{Test} 
& PerAct          & 94.03 & \textcolor{gray}{\textbf{97.76}} & 19.48 & \textcolor{gray}{24.68} & 24.64 & \textcolor{gray}{42.03} & 22.33 & \textcolor{gray}{45.63} \\
& PerAct†         & \textbf{94.78} & \textcolor{gray}{95.52} & 24.68 & \textcolor{gray}{28.57} & 23.19 & \textcolor{gray}{49.28} & 30.10 & \textcolor{gray}{48.54} \\
& PerAct (MT)†    & 90.30 & \textcolor{gray}{92.54} & 14.29 & \textcolor{gray}{20.78} & 20.29 & \textcolor{gray}{39.13} & 19.42 & \textcolor{gray}{37.86} \\
& ODYSSEY         & 60.45 & \textcolor{gray}{81.34} & \textbf{51.32} & \textcolor{gray}{\textbf{76.32}} & \textbf{56.30} & \textcolor{gray}{\textbf{64.70}} & \textbf{74.32} & \textcolor{gray}{\textbf{84.45}} \\
\midrule

\multirow{4}{*}{Novel Object}
& PerAct          & 86.55 & \textcolor{gray}{\textbf{92.73}} & 11.40 & \textcolor{gray}{35.09} & 0.00  & \textcolor{gray}{00.00} & 1.82  & \textcolor{gray}{5.45} \\
& PerAct†         & \textbf{87.27} & \textcolor{gray}{91.27} & 10.53 & \textcolor{gray}{32.46} & 0.00  & \textcolor{gray}{4.94} & 0.00  & \textcolor{gray}{5.45} \\
& PerAct (MT)†    & 81.09 & \textcolor{gray}{85.45} & 7.89  & \textcolor{gray}{24.56} & 0.00  & \textcolor{gray}{4.94} & 1.82  & \textcolor{gray}{5.45} \\
& ODYSSEY         & 41.09 & \textcolor{gray}{72.36} & \textbf{31.58} & \textcolor{gray}{\textbf{64.04}} & \textbf{50.32} & \textcolor{gray}{\textbf{61.29}} & \textbf{84.06} & \textcolor{gray}{\textbf{92.82}} \\
\midrule

\multirow{4}{*}{Novel Scene}
& PerAct          & \textbf{72.85} & \textcolor{gray}{\textbf{84.62}} & 17.07 & \textcolor{gray}{31.71} & 0.00  & \textcolor{gray}{4.97} & 5.10  & \textcolor{gray}{19.11} \\
& PerAct†         & 69.68 & \textcolor{gray}{84.16} & 13.41 & \textcolor{gray}{37.80} & 0.55  & \textcolor{gray}{4.97} & 6.37  & \textcolor{gray}{19.75} \\
& PerAct (MT)†    & 67.87 & \textcolor{gray}{79.64} & 7.32  & \textcolor{gray}{29.27} & 1.10  & \textcolor{gray}{4.42} & 7.01  & \textcolor{gray}{14.01} \\
& ODYSSEY         & 43.43 & \textcolor{gray}{63.80} & \textbf{41.98} & \textcolor{gray}{\textbf{54.32}} & \textbf{52.94} & \textcolor{gray}{\textbf{64.31}} & \textbf{62.96} & \textcolor{gray}{\textbf{70.37}} \\
\midrule

\multirow{4}{*}{Novel State}
& PerAct          & 2.38  & \textcolor{gray}{0.68} & 0.00  & \textcolor{gray}{0.95} & 0.00  & \textcolor{gray}{5.81} & 1.39  & \textcolor{gray}{1.39} \\
& PerAct†         & 0.68  & \textcolor{gray}{2.38} & 0.48  & \textcolor{gray}{00.00} & 0.00  & \textcolor{gray}{6.22} & 0.00  & \textcolor{gray}{8.33} \\
& PerAct (MT)†    & 2.04  & \textcolor{gray}{3.06} & 0.95  & \textcolor{gray}{2.38} & 0.00  & \textcolor{gray}{3.73} & 2.78  & \textcolor{gray}{11.11} \\
& ODYSSEY         & \textbf{46.60} & \textcolor{gray}{\textbf{78.91}} & \textbf{67.48} & \textcolor{gray}{\textbf{81.07}} & \textbf{48.85} & \textcolor{gray}{\textbf{63.22}} & \textbf{78.87} & \textcolor{gray}{\textbf{91.70}} \\
\midrule

\multirow{4}{*}{Any State}
& PerAct          & 47.01 & \textcolor{gray}{50.75} & 7.79  & \textcolor{gray}{19.48} & 5.80  & \textcolor{gray}{21.74} & 3.88  & \textcolor{gray}{15.53} \\
& PerAct†         & \textbf{48.51} & \textcolor{gray}{47.76} & 14.29 & \textcolor{gray}{14.29} & 4.35  & \textcolor{gray}{24.64} & 6.80  & \textcolor{gray}{13.59} \\
& PerAct (MT)†    & 46.27 & \textcolor{gray}{47.01} & 12.99 & \textcolor{gray}{12.99} & 4.35  & \textcolor{gray}{5.80} & 4.85  & \textcolor{gray}{8.74} \\
& ODYSSEY         & 47.22 & \textcolor{gray}{\textbf{80.59}} & \textbf{51.95} & \textcolor{gray}{\textbf{81.82}} & \textbf{54.78} & \textcolor{gray}{\textbf{67.83}} & \textbf{83.33} & \textcolor{gray}{\textbf{94.93}} \\
\bottomrule
\end{tabular}
\caption{\textbf{Evaluation results across short-horizon tasks and data splits.} Each row corresponds to a specific split, and each column reports success rates (\%) on a task. Gray entries denote phase-II performance using ground-truth grasp poses. ODYSSEY outperforms all PerAct variants, especially in generalization and manipulation settings.}
\label{tab:detailed-short-horizon-tasks}
\end{table*}

\begin{figure*}[h]
  \centering
  \includegraphics[width=1.0\linewidth]{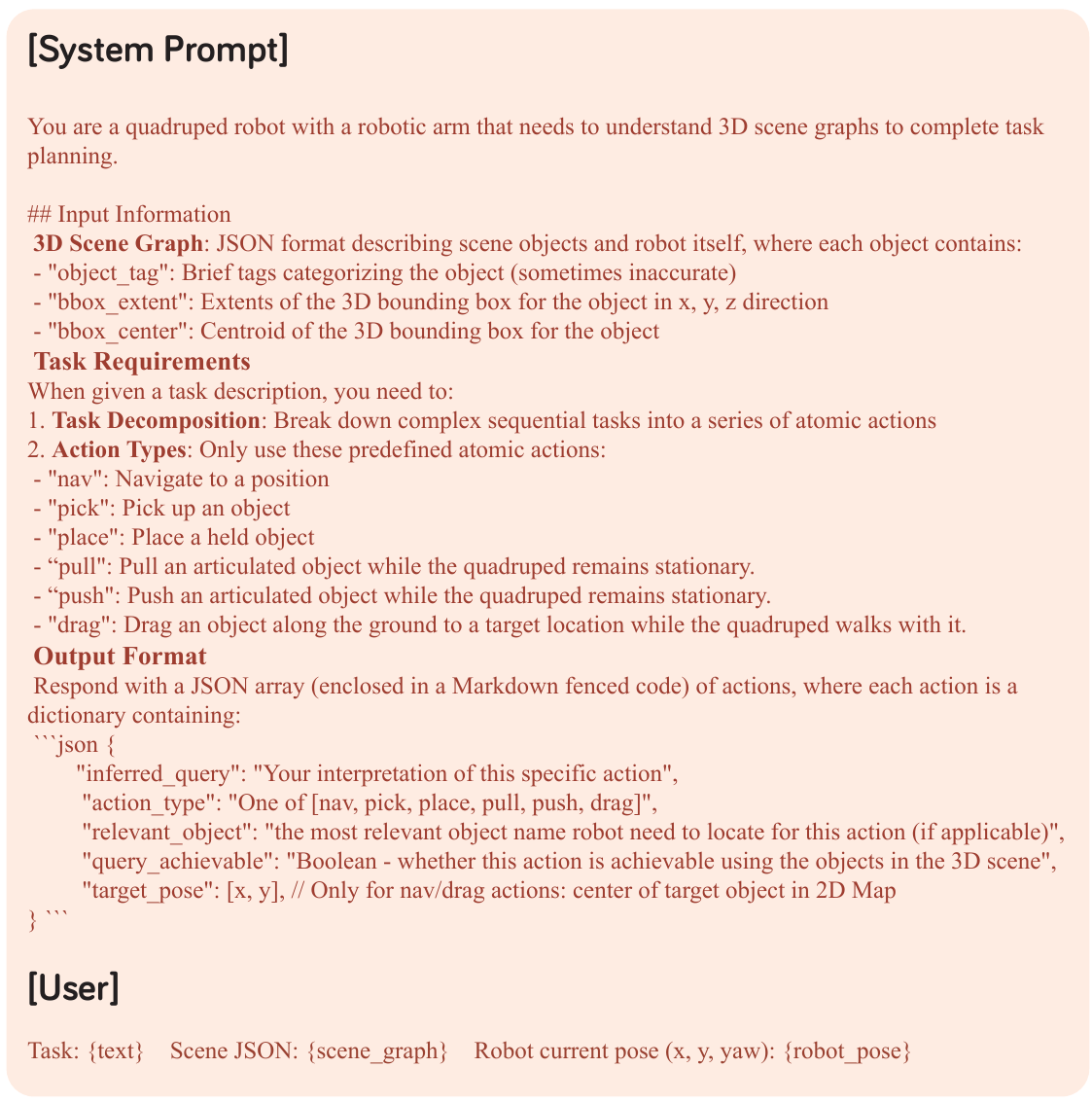}
  \caption{Prompt examples for task-level planning using LLM. The model is conditioned on a semantic instance graph and a goal instruction, and produces a structured list of atomic actions.}
  \label{fig:llm_prompt_examples}
\end{figure*}

\end{document}